%% file: deepthink_arxiv.tex
\documentclass[11pt,letterpaper]{article}

\input{Marcos/macro_deepthink}

\usepackage{amsmath,amssymb,amsthm}
\usepackage{deepthink}

\usepackage[nameinlink]{cleveref}
\usepackage[
  backend=biber,
  style=alphabetic,
  maxbibnames=100,
  minbibnames=100,
  maxcitenames=2,
  mincitenames=2
]{biblatex}
\newtheorem{innercustomthm}{Proposition}
\newenvironment{mythm}[1]
  {\renewcommand\theinnercustomthm{#1}\innercustomthm}
  {\endinnercustomthm}
  
\title{The Emergence of Reproducibility and Generalizability in Diffusion Models}

\newcommand{\jointfirst}{\textsuperscript{\dag}}
\newcommand{\corrauth}{\textsuperscript{\ddag}}

\authorblock{
  \href{https://www.huijiezh.com/}{Huijie Zhang}\jointfirst,
  \href{https://scholar.google.com/citations?user=O3Df4PwAAAAJ&hl=en}{Jinfan Zhou}\jointfirst,
  \href{https://scholar.google.com/citations?user=ybsmKpsAAAAJ&hl=en}{Yifu Lu},
  \href{https://www.linkedin.com/in/minzhe-guo/}{Minzhe Guo},
  \href{https://peng8wang.github.io/}{Peng Wang},
  \href{https://liyueshen.engin.umich.edu/}{Liyue Shen},
  \href{https://qingqu.engin.umich.edu/}{Qing Qu}\corrauth
}

\affiliation{
   University of Michigan
}

\authornote{
  \jointfirst\ Joint first author \quad
  \corrauth\ Corresponding author
}

\abstracttext{
\noindent In this work, we investigate an intriguing and prevalent phenomenon of diffusion models which we term as ``consistent model reproducibility'': given the same starting noise input and a deterministic sampler, different diffusion models often yield remarkably similar outputs. 
We confirm this phenomenon through comprehensive experiments, implying that different diffusion models consistently reach the same data distribution and score function regardless of diffusion model frameworks, model architectures, or training procedures. 
More strikingly, our further investigation implies that diffusion models are learning \emph{distinct distributions} affected by the training data size. 
This is supported by the fact that the model reproducibility manifests in two distinct training regimes: (i) ``memorization regime,'' where the diffusion model overfits to the training data distribution, and (ii) ``generalization regime,'' where the model learns the underlying data distribution. Our study also finds that this valuable property generalizes to many variants of diffusion models, including those for conditional generation, solving inverse problems, and model fine-tuning. Finally, our work raises numerous intriguing theoretical questions for future investigation and highlights practical implications regarding training efficiency, model privacy, and the controlled generation of diffusion models. 
}

\keywords{Diffusion Model, Reproducibility, Memorization and Generalization, Interpretability}

\date{\today}
\correspondence{\href{mailto:huijiezh@umich.edu}{huijiezh@umich.edu}}
\resources{\quad \href{https://www.huijiezh.com/reproducibility-generalizability/index.html}{Project page} \quad $\cdot$ \quad \href{https://github.com/huijieZH/Diffusion-Model-Generalizability}{Code}}

\headerlogo{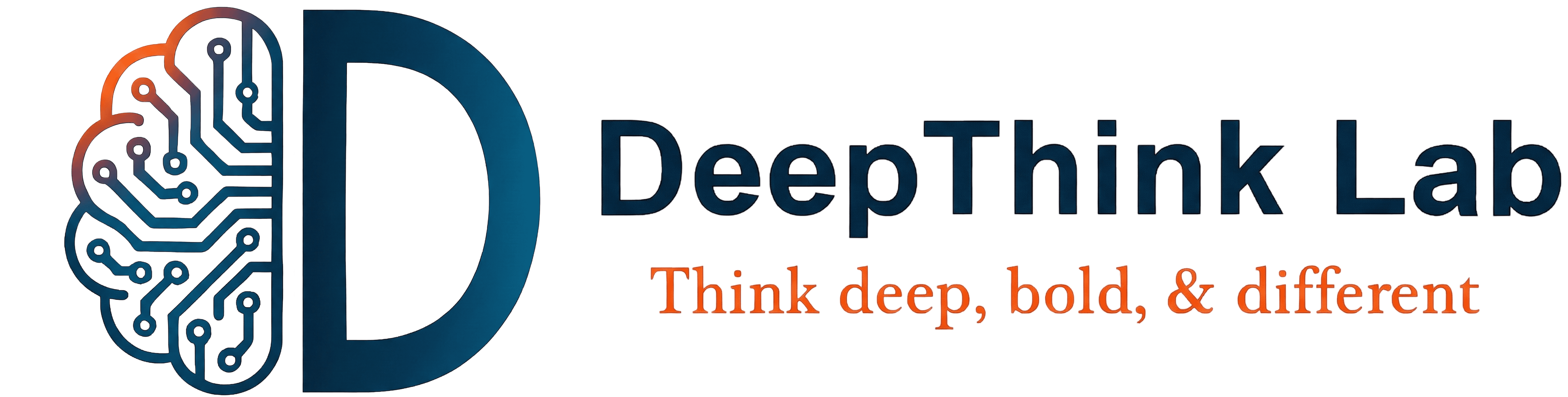}{https://deepthink-umich.github.io}

\begin{document}

\makeDeepthinkHeader
\vspace{-0.1in}
\begin{figure}[h]
     \centering
     \begin{subfigure}[t]{0.4\columnwidth}
         \centering
         \includegraphics[width=\linewidth]{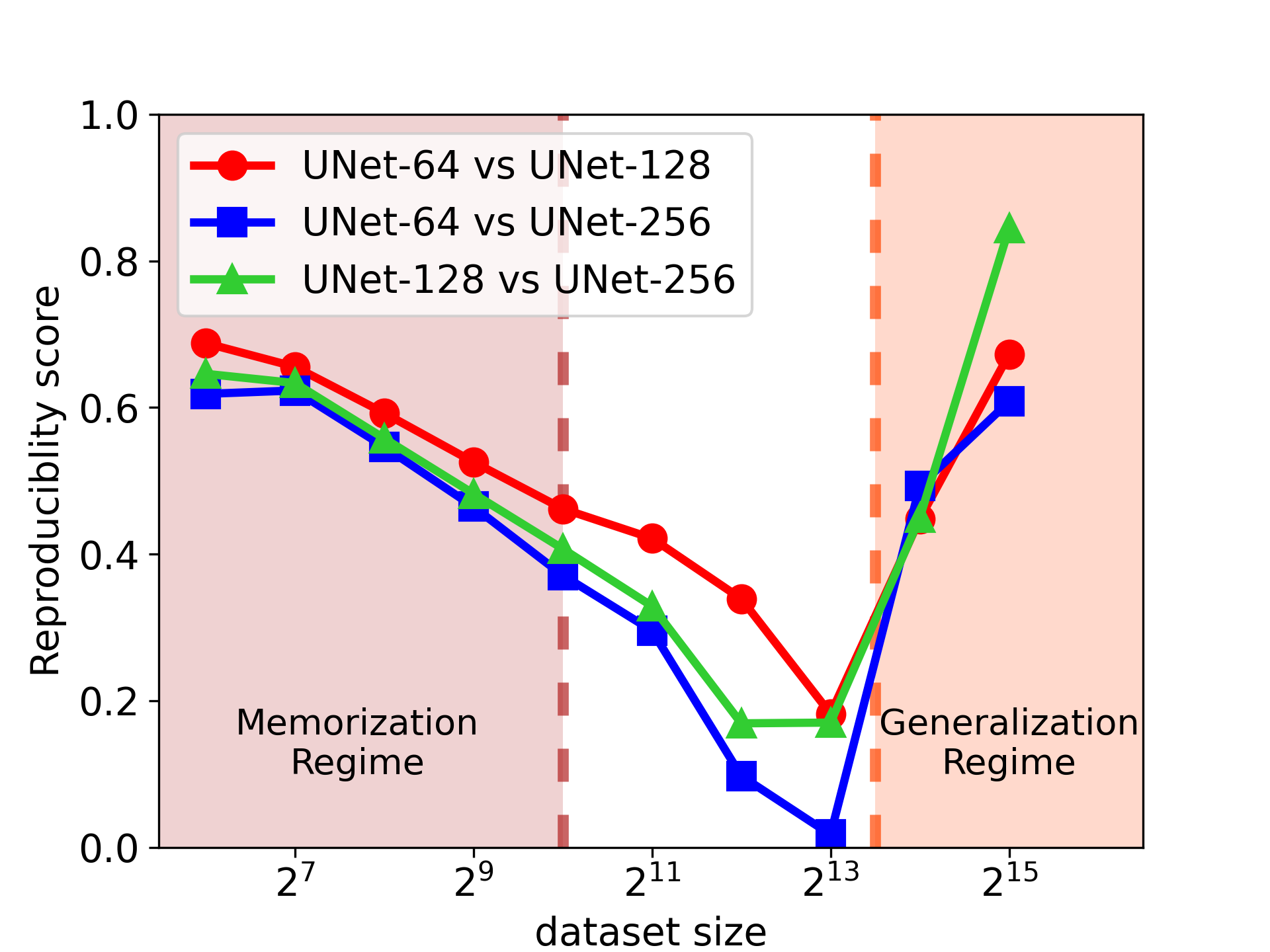}
         \caption{Reproducibilty}
         \label{fig:reproducibility}
     \end{subfigure}
     \begin{subfigure}[t]{0.4\columnwidth}
         \centering
         \includegraphics[width=\linewidth]{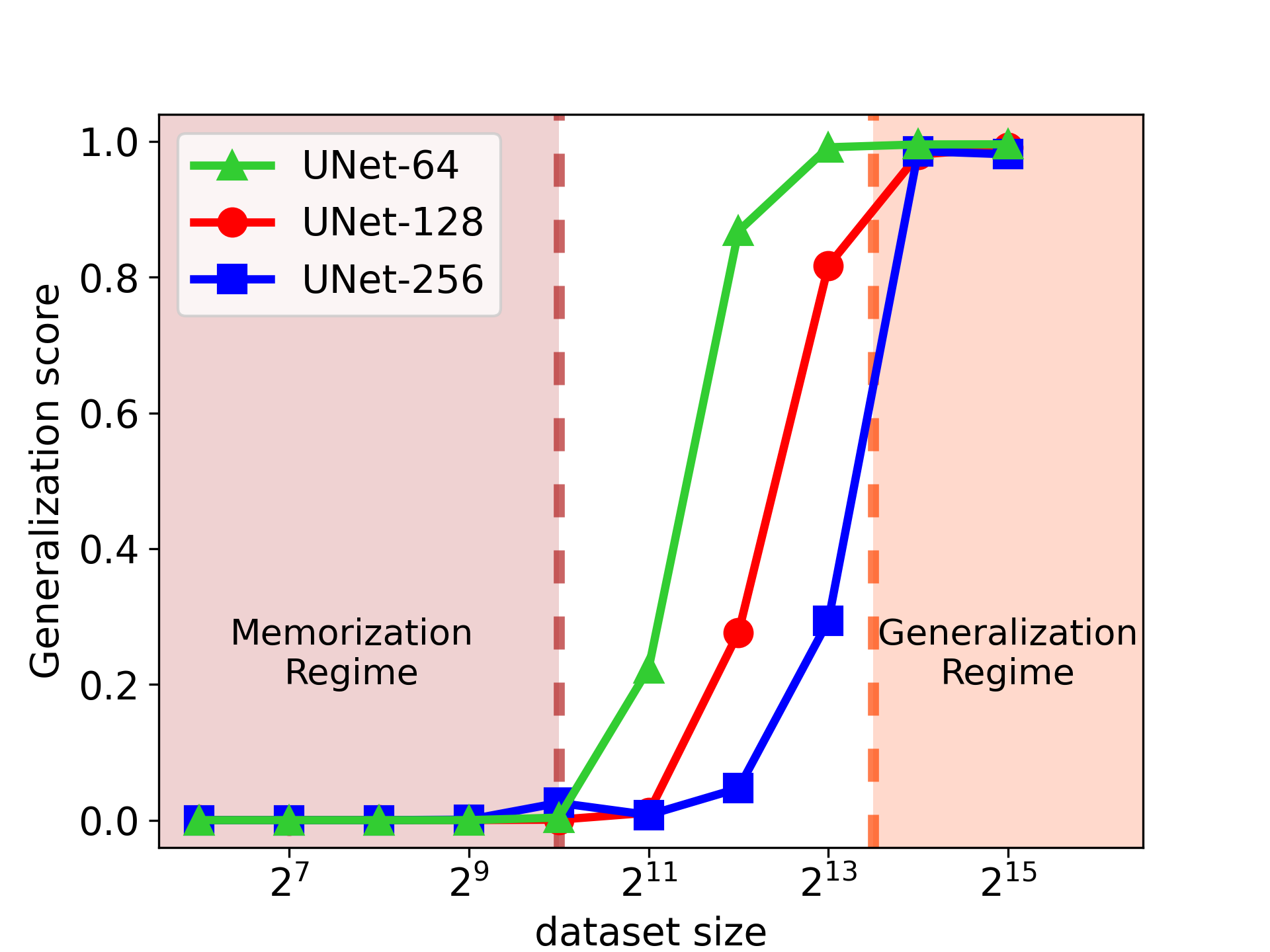}
         \caption{Generalizability}
         \label{fig:regime}
     \end{subfigure}
     \hfill
     \caption{\textbf{``Memorization'' and ``Generalization'' regimes for unconditional diffusion models.} 
     We employ DDPMs trained on the CIFAR-10 dataset, varying both the model capacity and the size of the training dataset.
     The figure on the left displays the reproducibility score as we compare various models across different dataset sizes, while the figure on the right illustrates the generalizability score of the models as the dataset size changes.
     }    
   \label{fig:two_regime}
\end{figure}


\newpage
\tableofcontents
\newpage

\input{section_new/1.intro}

\input{section_new/2.phenomenom_regimes}

\input{section_new/3.analysis_tworegime}

\input{section_new/4.phenomenon_extension}
\input{section_new/5.related_works}

\input{section_new/6.discussion}

\section*{Acknowledgement}
HJZ, YFL, PW, and QQ acknowledge support from NSF CAREER CCF-2143904, NSF CCF-2212066, NSF CCF-2212326, NSF IIS 2312842, ONR N00014-22-1-2529, an AWS AI Award, and a gift grant from KLA. LYS and QQ acknowledge support from MICDE Catalyst Grant, and LYS also acknowledges the support from the MIDAS PODS Grant. Results presented in this paper were obtained using CloudBank, which is supported by the NSF under Award \#1925001, and the authors acknowledge efficient cloud management framework SkyPilot \cite{yang2023skypilot} for computing. The authors acknowledge valuable discussions with Prof. Jeffrey Fessler (U. Michigan), Prof. Saiprasad Ravishankar (MSU), Prof. Rongrong Wang (MSU), Prof. Weijie Su (Upenn), Dr. Ruiqi Gao (Google DeepMind), Mr. Bowen Song (U. Michigan), Mr. Xiao Li (U. Michigan), Mr. Zekai Zhang (U. Tsinghua), Dr. Ismail R. Alkhouri (U. Michigan and MSU)

\section*{Impact Statement}
This paper presents work whose goal is to advance the field of Machine Learning. On potential negative social impact is discussed in \Cref{sec:conclusion}. Given the reproducibility, commercial black-box diffusion models are susceptible to replication, adversarial attacks, and leaks of training data.

\newpage



\printbibliography
\newpage 

\appendices

We include more comprehensive experiment settings, quantitative results, and detailed discussion of the unconditional diffusion model in \Cref{append:unconditional}, 
theoretical proof in \Cref{append:theory}, experiment setting for \Cref{sec:analysis_generalization_gt} in \Cref{sec:analysis_generalization_exp_setting}, experiment settings for conditional diffusion model in \Cref{append:conditional}, stable diffusion in \Cref{append:stablediffusion}, diffusion model for solving inverse problems in \Cref{append:inverseproblem}, fine-tuning diffusion model in \Cref{append:fintuning}.

\input{section_new/appendix/Appendix_unconditional}

\input{section_new/appendix/Appendix_theory}
\input{section_new/appendix/Appendix_distrib_learning}
\input{section_new/appendix/Appendix_conditional}

\input{section_new/appendix/Appendix_stablediffusion}

\input{section_new/appendix/Appendix_inverseproblem}
\input{section_new/appendix/Appendix_finetunning}

\end{document}

%% file: Marcos/macro_deepthink.tex
\usepackage{tgpagella}
\usepackage[utf8]{inputenc} 
\usepackage[T1]{fontenc}    
\usepackage{url}
\usepackage[hidelinks]{hyperref}
\usepackage{amsmath,amsthm,amssymb,amsbsy,amsfonts,amscd,bm}
\usepackage{paralist}
\usepackage{xcolor}
\usepackage{color}
\usepackage{graphicx}
\graphicspath{{./figs/}}
\usepackage{algorithm}
\usepackage{comment}
\usepackage{multirow}
\usepackage{enumitem}
\usepackage{fancyhdr}
\usepackage{authblk}
\usepackage{subcaption}
\usepackage{wrapfig}
\usepackage{bbding}
\usepackage{algpseudocode}
\usepackage{graphicx}
\usepackage{bbold}
\usepackage{mathtools}
\usepackage{multirow}
\usepackage{enumitem}
\usepackage{bbding}
\usepackage{subcaption}
\usepackage{wrapfig}
\usepackage{titletoc}
\usepackage{titlesec}
\usepackage{amsthm}

\usepackage[toc, page]{appendix}

\newtheorem{lemma}{Lemma}

\theoremstyle{remark}

\newcommand{\R}{\mathbb{R}}

\newcommand{\e}{\begin{equation}}
\newcommand{\ee}{\end{equation}}
\newcommand{\en}{\begin{equation*}}
\newcommand{\een}{\end{equation*}}
\newcommand{\eqn}{\begin{eqnarray}}
\newcommand{\eeqn}{\end{eqnarray}}
\newcommand{\bmat}{\begin{bmatrix}}
\newcommand{\emat}{\end{bmatrix}}

\DeclareMathAlphabet\mathbfcal{OMS}{cmsy}{b}{n}

\newcommand{\E}{\operatorname{\mathbb{E}}}

\DeclareMathOperator*{\argmax}{\text{arg~max}}

\newcommand{\eps}{\epsilon}

\hypersetup{
    colorlinks=true,%
    citecolor=blue,%
    filecolor=blue,%
    linkcolor=blue,%
    urlcolor=blue
}

 \newcommand{ \Brac }[1]{\left\lbrace #1 \right\rbrace}
 \newcommand{ \brac }[1]{\left[ #1 \right]}

\graphicspath{{./figs/}}

\newlength{\imgwidth}
\newboolean{twoColVersion}
\setboolean{twoColVersion}{false}
\newcommand{\twoCol}[2]{\ifthenelse{\boolean{twoColVersion}} {#1} {#2} }

\newtheorem{proposition}{\bf{Proposition}}

\long\def\comment#1{}

\def\E{\mathop{\rm E\,\!}\nolimits}

\newcommand{\paren}[1]{\left( #1 \right)}
\long\def\red#1{\bgroup\color{red}#1\egroup}

\definecolor{mich-blue}{HTML}{0027CC}
\definecolor{mich-blue-high}{HTML}{0027CC}
\definecolor{red-high}{HTML}{CA2020}
\definecolor{green-high}{HTML}{20A520}
\definecolor{mich-maize}{HTML}{FFCB05}
\definecolor{law-stone}{HTML}{655A52}
\definecolor{burton-beige}{HTML}{9B9A9D}
\definecolor{arch-ivy}{HTML}{7E732F}

 \colorlet{color1}{gray!15}

%% file: section_new/1.intro.tex
\section{Introduction}

 Recently, diffusion models have emerged as a powerful new family of deep generative models with remarkable performance in many applications, including image generation \cite{ho2020denoising, song2020score, rombach2022high}
 , image-to-image translation \cite{su2022dual,saharia2022palette, zhao2022egsde}, text-to-image synthesis \cite{rombach2022high, ramesh2021zero, nichol2021glide}, and solving inverse problem \cite{chung2022improving, song2022pseudoinverse, chung2022diffusion,song2023solving}.
 These models learn an unknown data distribution generated from the Gaussian noise distribution through a process that imitates the non-equilibrium thermodynamic diffusion process \cite{ho2020denoising, song2020score}. In the forward diffusion process, the noise is continuously injected into training samples; while in the reverse diffusion process, a model is learned to remove the noise from noisy samples parametrized by a noise-predictor neural network. Then guided by the trained model, new samples (e.g., images) from the target data distribution can be generated by transforming random noise instances through step-by-step denoising following the reverse diffusion process. Despite the remarkable data generation capabilities, the fundamental mechanisms driving their performance are largely under-explored.

\begin{figure}[t]
     \centering
     \includegraphics[width=.8\columnwidth]{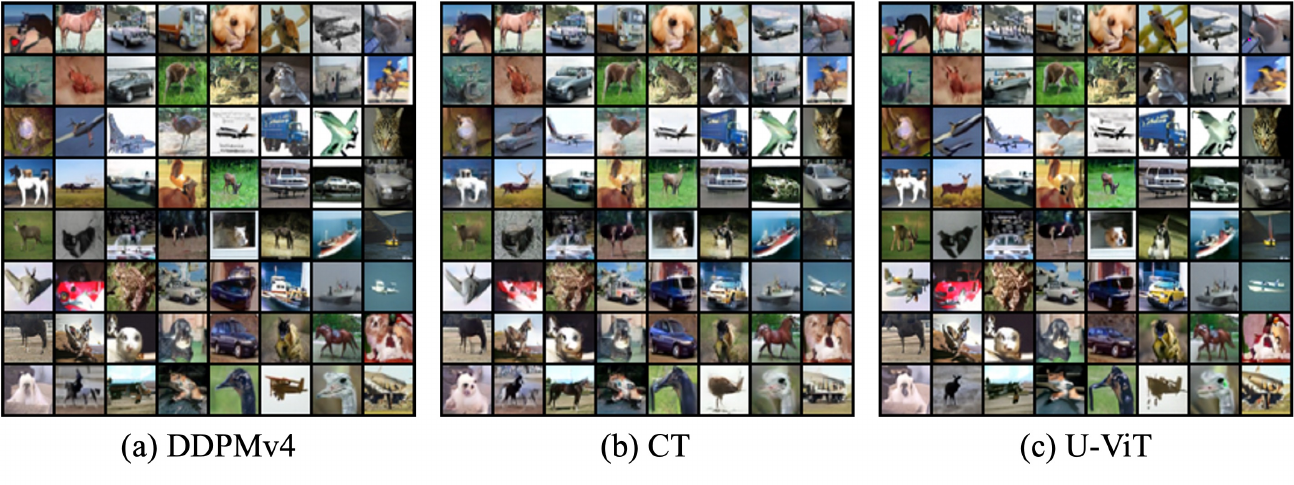}
     \caption{\textbf{Visualization of generation samples from different diffusion models.} We utilized denoising diffusion probabilistic models (DDPM) \cite{ho2020denoising, song2020denoising}, consistency model (CT) \cite{song2023consistency}, U-ViT \cite{bao2023all} trained on CIFAR-10 \cite{krizhevsky2009learning} dataset. Samples in the corresponding row and column are generated from the same initial noise with a deterministic ODE sampler. 
     }
     \label{fig:unconditional_sample}
\end{figure}



\renewcommand\thempfootnote{\arabic{mpfootnote}}
In this work, we study an intriguing while prevalent phenomenon that sets diffusion models apart from most other generative models. We refer to this phenomenon as ``\emph{consistent model reproducibility}''. More precisely, as illustrated in \Cref{fig:unconditional_sample}, when different diffusion models are trained on the same dataset, and sampled from the \emph{same} noises when using a deterministic ODE sampler.\footnote{We employ a deterministic sampler to ensure model reproducibility, but stochastic samplers can also achieve reproducibility when they generate consistent noise across different models.}
\begin{tcolorbox}
\begin{center}
\emph{Different diffusion models consistently converge to \textbf{nearly identical} image contents, which is \emph{irrespective} of network architectures, training and sampling procedures, and perturbation kernels.}
\end{center}
\end{tcolorbox}
This phenomenon implies that different diffusion models are learning nearly identical mapping and distributions, as further discussed in \Cref{sec:analysis_tworegimes}. More interestingly, through studying the reproducibility under different regimes of training data size, we further find that diffusion models are learning \emph{different types} of data distributions depending on the size of training data. As illustrated in \Cref{fig:two_regime}, this is corroborated by our findings that the consistent model reproducibility emerges in two distinct regimes: 
(\emph{i}) ``\textbf{\emph{Memorization regime}}'': the model has the capacity to memorize the training data but no ability to generate new samples. The co-existence of reproducibility and memorization implies that the diffusion model is learning the empirical multi-delta distribution of the training samples. (\emph{ii}) ``\textbf{\emph{Generalization regime}}'': the model regains reproducibility while it gains the ability to produce new data. The co-emergence of reproducibility and generalizability indicates that the diffusion model is learning the underlying distribution of the data.

\paragraph{Summary of contributions.} In summary, we briefly highlight our contributions below:
\begin{itemize}[leftmargin=*]
    \itemsep0em
    \item \textbf{A comprehensive study of model reproducibility.} We present the first comprehensive and systematic study of the reproducibility in diffusion models. Our findings are consistent under various network architectures, noise perturbation kernels, training and sampling settings.
    \item \textbf{Two regimes of model reproducibility and distribution learning.} Our analysis reveals that reproducibility manifests in two regimes. We demonstrate that diffusion models learn different types of distributions (i.e., empirical vs. underlying distribution) in different regimes. 
    \item \textbf{Model reproducibility beyond unconditional diffusion models.} Under various different settings, we show that reproducibility manifest in different but structured ways, including conditional diffusion models, inverse problem solving, fine-tuning.
\end{itemize}

\paragraph{Theoretical and practical implications of our work.} Understanding model reproducibility within diffusion models could carry significant implications for both theoretical and practical aspects. Theoretically, understanding the question will shed light on how the mapping function between the noise and data distributions is learned and constructed, and it will also offer profound understanding of how diffusion models are capable of learning the complicated image distribution from a limited number of training samples. We discuss the theoretical aspects in more detail in \Cref{sec:related}.
In practical terms, gaining a deeper insight into model reproducibility could potentially lead to (1) improved efficiency in training, (2) solutions for data privacy concerns in large-scale pre-trained diffusion models, and (3) more interpretable and controllable data generation processes.  We further discuss the practical aspect in detail in \Cref{sec:conclusion}. 




\paragraph{Notations.} We denote scalar (function) as regular lower-case letters (e.g. $s_t, f(t)$), vectors (function) with bold lower-case letters (e.g. $\bm x, \bm s(\bm x_t, t)$). We use $[N]$ to denote the set $\{1, 2, \ldots, N\}$,  $\mathbb P(\cdot)$ to denote the probability, $\mathbb E\left[\cdot\right]$ to denote expectation, $||\cdot||_2$ to denote L2 norm, $\mathcal N(\cdot)$ to denote gaussian distribution, $\mathcal U(0, 1)$ to denote uniform distribution from 0 to 1. Given any $d \in \mathbb N$, we use $\bm I_d$ to denote an identity matrix of size $d$.

\paragraph{Organization of the paper.}

The rest of the paper is organized as follows: \Cref{sec:tworegime} introduces and \Cref{sec:analysis_tworegimes} analyzes the reproducibility in the contexts of memorization and generalization regimes. We then broaden our investigation to include variants of diffusion model settings in \Cref{sec:reproducibility_more}. \Cref{sec:related} draws comparisons between our work and related literature. Finally, in \Cref{sec:conclusion}, we explore the implications of our empirical findings.

%% file: section_new/2.phenomenom_regimes.tex
\section{Consistent Model Reproducibility}
\label{sec:tworegime}


While the illustrations in \Cref{fig:unconditional_sample} and initial investigations in the seminal work \cite{song2020score} are motivating, this work provides a more comprehensive and systematic study of model reproducibility in diffusion models.\footnote{Recent seminal work \cite{song2020score} has observed a similar phenomenon (see also subsequent works \cite{song2023consistency, karras2022elucidating}), but the study in \cite{song2020score} remains preliminary.} We begin by proposing quantitative metrics to evaluate reproducibility as well as generalizability in diffusion models. Subsequently, we discover a strong relationship between the reproducibility and generalizability of diffusion models.


\subsection{Measures of Reproducibility and Generalizability}\label{sec:metric}

\paragraph{Measure of model reproducbility.} 
To study the reproducibility phenomenon in \Cref{fig:unconditional_sample} more quantitatively, we introduce the \textit{reproducibility (RP) score} to measure the similarity of image pair generated from two different diffusion models starting from the \emph{same noise}, which is drawn \emph{i.i.d.} from the standard Gaussian distribution: 
\begin{align*}
    \text{RP Score} \;:=\; \mathbb{P}\paren{\mathcal{M}_{\text{SSCD}}(\bm x_1, \bm x_2)>0.6},
\end{align*}
which measures the \emph{probability} of a generated sample pair $(\bm x_1, \bm x_2)$ from two different diffusion models to have \emph{self-supervised copy detection} (SSCD) similarity $\mathcal{M}_{\text{SSCD}}$ larger than $0.6$ \cite{pizzi2022self, somepalli2023understanding}.\footnote{As demonstrated in \cite{somepalli2023understanding}, $\mathcal{M}_{\text{SSCD}} > 0.4$ already exhibits very strong visual similarities.} Higher RP score indicates stronger model reproducibility. In practice, we estimate \emph{RP Score} by the empirical probability using 10K noise samples.
The SSCD similarity is first introduced in \cite{pizzi2022self} to measure the replication between image pair $(\bm x_1, \bm x_2)$, which is defined as follows: 
\begin{align*}
    \mathcal{M}_{\text{SSCD}}(\bm x_1, \bm x_2) = \dfrac{\text{SSCD}(\bm x_1) \cdot \text{SSCD}(\bm x_2)}{||\text{SSCD}(\bm x_1)||_2 \cdot ||\text{SSCD}(\bm x_2)||_2}
\end{align*}
where $\text{SSCD}(\cdot)$ represents a neural descriptor for copy detection of images. 

In addition, we also use the \emph{mean-absolute-error (MAE) score} to measure the reproducibility, $\text{MAE Score} := \mathbb{P}\paren{\text{MAE}(\bm x_1, \bm x_2)<15.0}$, based upon similar setting with the RP score. $\text{MAE}(\cdot)$ is the operator that measures the mean absolute different of image pairs in the pixel value space ([0, 255]).

\paragraph{Measure of model generalizability.} Moreover, we discover a strong relationship between model reproducibility and its generalizability, where the latter refers to the model's ability to produce \emph{new samples} distinct from the ones in the training set. To assess the generalizability of diffusion models, we introduce the \textit{generalization (GL) score} as follows:
\begin{align*}
    \text{GL Score} :=1 - \mathbb{P}\paren{\max_{i\in[N]} \left[\mathcal{M}_{\text{SSCD}}(\bm x, \bm y_i)\right]>0.6},
\end{align*}
which is defined based upon the \emph{probability} of maximum $\mathcal{M}_{\text{SSCD}}$ over the training dataset larger than $0.6$. Similar to RP score, we empirically sample 10K initial noises to estimate the probability. Intuitively, GL score measures the dissimilarity between the generated sample $\bm x$ and all $N$ samples $\bm y_i$ from the training dataset $\Brac{\bm y_i }_{i=1}^N $. Higher GL score indicates stronger generalizability.



\subsection{Model Reproducibility Manifests in Two Regimes}\label{subsec:reproducibility-gen}

Based upon RP and MAE scores, we provide comprehensive quantitative studies (see \Cref{fig:reproducibility_selected}) to demonstrate the prevalence of model reproducibility in diffusion models. More interestingly, we discover that the reproducibility of the model arises either through memorization of the training data or by acquiring the ability to generalize. As highlighted in \Cref{fig:two_regime}, we show that 
\begin{tcolorbox}
\begin{center}
\emph{The model reproducibility manifests in two distinct \textbf{memorization} and \textbf{generalization} regimes, depending on the size of training data and model capacities.}
\end{center}
\end{tcolorbox}
In the following, we discuss the two regimes in detail:
\begin{itemize}[leftmargin=*]
    \item \textbf{``Memorization regime''} characterizes the scenario where the reproducibility is due to the memorization of the training data distribution. As illustrated in the left region of \Cref{fig:reproducibility}, this regime occurs when the model has much larger capacity than the size of training data. Although the model possesses the ability to reproduce the same results starting from the same noise, the generated samples are only replications of the samples in the training data and the model lacks the ability to generate new samples; see the left region of \Cref{fig:regime}. In this regime, the emergence of reproducibility is due to the fact that all diffusion models memorize the same multi-delta distribution of training samples. This can be verified by characterizing the closed-form solution of the score function under empirical multi-delta distribution (see \Cref{proposition:empirical distribution}), and by showing that practical diffusion models converge to such score function (see \Cref{fig:theoretical_verification}). An in-depth study is provided in \Cref{sec:analysis_memorization}. It should noted that, given no generalizability, training diffusion models in this regime might hold limited practical interest. 
    \item \textbf{``Generalization regime''} emerges when the diffusion model not only regains its reproducibility but also becomes capable of generating new samples distinct from the training data; see the right region of \Cref{fig:regime}. This usually happens when the diffusion model is trained on large dataset without full capacity to memorize the whole dataset \cite{yoon2023diffusion}; see the right region of \Cref{fig:reproducibility}. This is the regime in which diffusion models are commonly trained and employed in practice. As illustrated in \Cref{fig:regime}, we revealed that there is a clear \emph{phase transition} from the memorization regime to the generalization regime as the training samples increase. In the generalization regime, the model reproducbility co-emerges with the model's generalizability. We believe this is because all diffusion models are learning the same score function of the true underlying data distribution instead of the training data distribution. We provide an in-depth study in \Cref{sec:analysis_generalization}.
\end{itemize}

\subsection{Reproducibility is Rare in Generative Models}

\begin{figure}[t]
    \centering
    \includegraphics[width=0.5\linewidth]{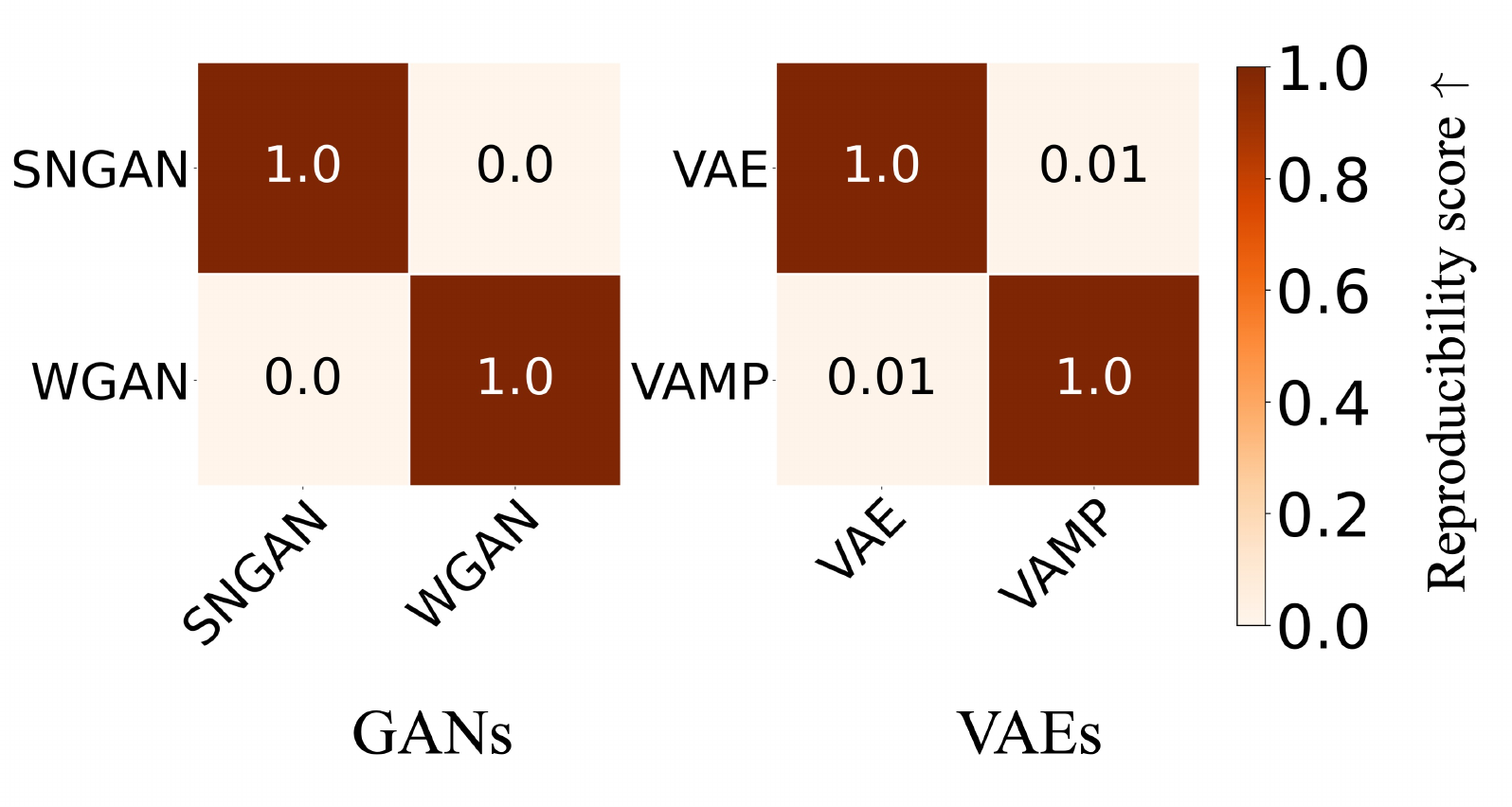}
    \caption{\textbf{Quantitative results for GANS and VAEs.} In our evaluation of GAN-based methods, we utilize two architectures: Wasserstein GAN (wGAN) \cite{arjovsky2017wasserstein} and Spectral Normalization GAN (SNGAN) \cite{miyato2018spectral} training on the CIFAR-10 dataset. For VAE-based approaches, we consider both the standard VAE and the Variational Autoencoding Mutual Information Bottleneck (VAMP) model \cite{tomczak2018vae} training on the MNIST \cite{lecun1998gradient} dataset.}
    \label{fig:gan_and_vae}
\end{figure}
We end this section by highlighting that only diffusion models appear to consistently exhibit model reproducibility. This property rarely exists in other generative models, with one exception as noted in \cite{khemakhem2020variational}.\footnote{\cite{khemakhem2020variational} demonstrates that VAE is uniquely identifiable encoding given a factorized prior distribution over the latent variables.} Quantitative results of model reproducibility for Generative Adversarial Network (GAN) \cite{goodfellow2014generative} and Variational Autoencoder (VAE) \cite{kingma2013auto} are in \Cref{fig:gan_and_vae}. 
In contrast to diffusion models, the observed lack of reproducibility in GANs and VAEs implies that they are not effectively trained to capture the underlying data distribution. This deficiency is a contributing factor to the occurrence of mode collapse in GANs \cite{arora2017gans}.


%% file: section_new/3.analysis_tworegime.tex
\section{Analyzing Reproducibility in Two Regimes} \label{sec:analysis_tworegimes}

To understand why diffusion models exhibit reproducibility across different models and regimes, it would be intuitive to first look at the reverse sampling process. When we employ an ODE sampler \cite{song2020score}, for $t \in [0,1]$ the reverse sampling process can be characterized by
\begin{align}\label{eqn:ODE-sampler}
    \bm x_t =  \paren{1- f(t') } \bm x_{t'}  + \frac{g^2(t')}{2} \cdot \underset{\textbf{score function}}{\bm s(\bm x_{t'};t')},
\end{align}
where $t' = t + \Delta t \in [0,1]$, $f(t) = \dfrac{\text{d} \log s_t}{\text{d} t}$, and $g^2(t) = \dfrac{\text{d} s^2_t \sigma^2_t}{\text{d} t} - 2 s^2_t \sigma^2_t \dfrac{\text{d} \log s_t}{\text{d} t}$. Here, the scalars $\sigma_t$ and $s_t$ denote the parameters of the perturbation kernel $p_{t}(\bm x_t|\bm x_0) = \mathcal{N}(\bm x_t;s_t\bm x_0, s_t^2\sigma_t^2\textbf{I})$ at the $t$-th time-step. Furthermore, let $f_{\bm{s}}: \mathcal E \mapsto \mathcal I$ be the mapping from the noise space $\mathcal E$ to the image space $\mathcal I$, by using a deterministic ODE sampler and the score function $\bm{s}(\bm x_t;t)$. The reproducibility of diffusion models is the result of the learned mapping $f_{\bm{s}}$ is reproducible.

Observing \eqref{eqn:ODE-sampler}, it becomes evident that the behavior of the ODE sampler is deterministic, with results exclusively reliant on $\bm s(\bm x_t;t)$. This implies that the consistency observed across various diffusion models might be attributed to the reproducibility in score matching. Hence, to understand the reproducibility observed in both memorization and generalization regimes as outlined in \Cref{sec:tworegime}, we must delve into two critical questions:
\begin{itemize}[leftmargin=*]
    \item \emph{How well do diffusion models approximate the score function $\bm{s}(\bm x_t;t)$ in each regime?}
    \item \emph{For each regime, which distribution $p(\bm x_0)$ do diffusion models learn the score function $\bm{s}(\bm x_t;t)$ from?}
\end{itemize}
In the following, we study both questions for the memorization and generalization regimes in \Cref{sec:analysis_memorization} and \Cref{sec:analysis_generalization}, respectively. Before that, we first derive the analytical form of any given distribution $p(\bm x_0)$ based upon the Tweedie's formula \cite{luo2022understanding} as follows.
\begin{lemma}\label{lem:key}
   Suppose the distribution learned by diffusion model is $p(\bm x_0)$ and the perturbation kernel $p_{t}(\bm x_t|\bm x_0) = \mathcal{N}(\bm x_t;s_t\bm x_0, s_t^2\sigma_t^2\textbf{I})$ with perturbation parameters $s_t, \sigma_t$. The ideal score function has the following form
    \begin{align*} 
    \begin{split}
    &\bm{s}(\bm x_t;t)
    = \frac{1}{s_t^2\sigma_t^2}\left(\E_{\bm x_t \sim p_{t}(\bm x_t)}[\bm x_0|\bm x_t] - \bm x_t\right)\\
    &= \frac{1}{s_t^2\sigma_t^2}\left(s_t\frac{\E_{\bm x_0 \sim p(\bm x_0)}[\mathcal{N}(\bm x_t;s_t\bm x_0, s_t^2\sigma_t^2\textbf{I}) \cdot \bm x_0]}{\E_{\bm x_0 \sim p(\bm x_0)}[\mathcal{N}(\bm x_t;s_t\bm x_0, s_t^2\sigma_t^2\textbf{I})]} - \bm x_t\right).
    \end{split}
    \end{align*}   
\end{lemma}
In the following, we will use the above result to derive the optimal score function w.r.t. different $p(\bm x_0)$ in two distinct regimes.

\subsection{Reproducibility in Memorization Regime} \label{sec:analysis_memorization}

\begin{figure}[t]
     \centering
     \includegraphics[width=\columnwidth]{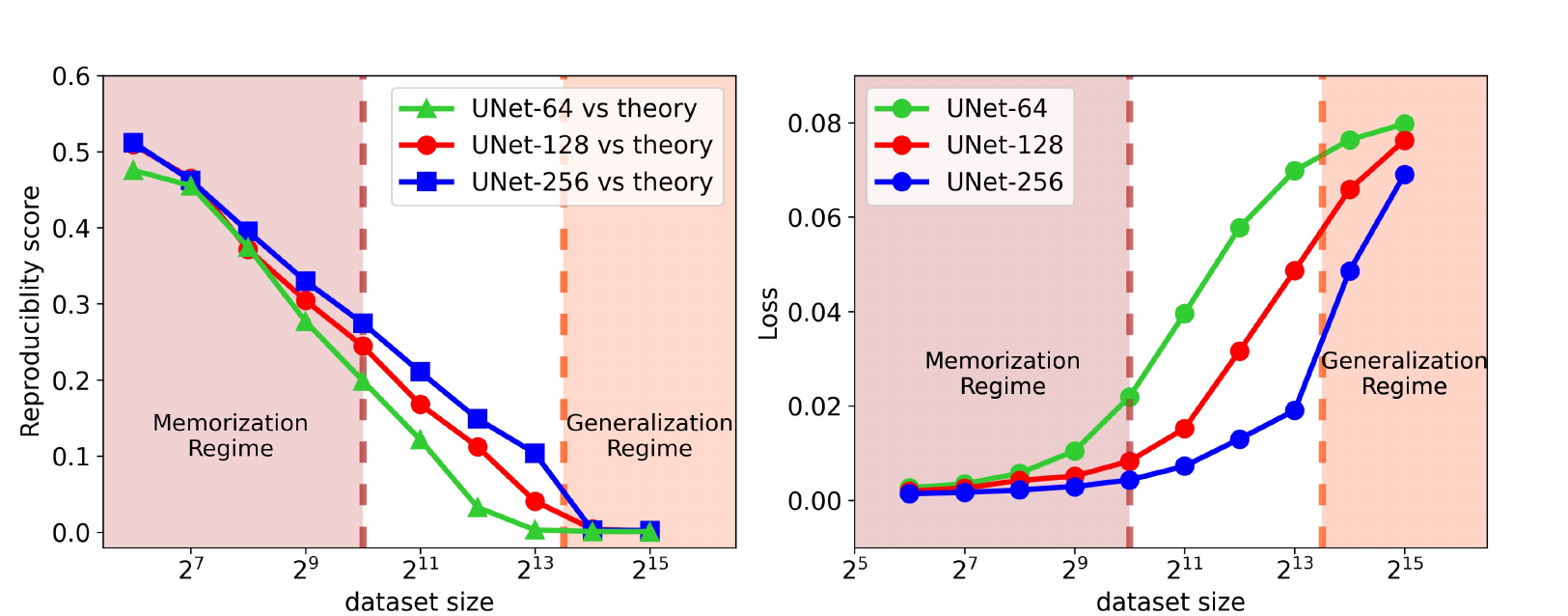}
     \caption{\textbf{Convergence of the optimal denoiser (left) and training loss (right) w.r.t. the training data size.} We employ DDPMv4 and conduct training on the CIFAR-10 dataset. During this process, we make modifications to both the model's capacity and the size of the training dataset, maintaining the same configuration as depicted in \Cref{fig:two_regime}. The left figure illustrates the reproducibility score between each diffusion model and the theoretically unique identifiable encoding as outlined in \Cref{proposition:empirical distribution}, the right figure illustrates the training loss for these models when trained till converge.}
    \label{fig:theoretical_verification}
\end{figure}

Through a combination of theoretical and experimental studies, we show that in the memorization regime,
\begin{tcolorbox}
   \emph{reproducibility is a result of memorizing the \textbf{training distribution} $p(\bm x_0) = \dfrac{1}{N} \sum_{i =1}^{N} \delta (\bm x_0 - \bm y_i)$.}
\end{tcolorbox}
Here, $p(\bm x_0)$ denotes the multi-delta distribution of the training samples $\Brac{\bm y_i }_{i=1}^N $ and $\delta(\cdot)$ denotes the Dirac delta function. In the following, we corroborate our claim by (i) deriving the optimal score function of $p(\bm x_0)$ in \Cref{proposition:empirical distribution}, and by (ii) showing that practical diffusion models converge to the optimal score function in the small data regime; see \Cref{fig:theoretical_verification}.


\begin{proposition}\label{proposition:empirical distribution}
   Given a training dataset $\Brac{\bm y_i }_{i=1}^N $ of $N$-samples, consider the same setting of \Cref{lem:key} with $p(\bm x_0)$ following the empirical multi-delta distribution $p(\bm x_0) = \frac{1}{N} \sum_{i =1}^{N} \delta (\bm x_0 - \bm y_i)$. In this setting, we can show that the score function can be characterized as 
    \begin{align*} 
    \begin{split}
        &\bm{s}_{\text{emp}}(\bm x_t;t) = -\frac{1}{s^2_t \sigma^2_t}\brac{\bm x_t - s_t\frac{\sum_{i = 1}^{N}\mathcal{N}(\bm x_t;s_t\bm y_i, s_t^2\sigma_t^2\textbf{I})\bm y_i}{\sum_{i = 1}^{N}\mathcal{N}(\bm x_t;s_t\bm y_i, s_t^2\sigma_t^2\textbf{I})}}.
    \end{split}
    \end{align*}
\end{proposition}
The proof for \Cref{proposition:empirical distribution} can be found in the \Cref{append:theory}, building upon previous findings from \cite{karras2022elucidating, yi2023generalization}. From \Cref{proposition:empirical distribution}, we can see that the score function $ \bm{s}_{\text{emp}}(\bm x_t;t)$ is purely determined by the given training dataset $\Brac{\bm y_i }_{i=1}^N $ and perturbation parameters $s_t, \sigma_t$. 

Moreover, by comparing the reproducibility between the theoretical noise-to-image mapping $f_{\bm{s}_{\text{emp}}}$ and different practically trained diffusion models, our experiments in \Cref{fig:theoretical_verification} (left) demonstrate that 
the trained networks have a very \emph{high similarity} compared with the theoretical solution when the training data size is small enough.
In the meanwhile, the training loss in \Cref{fig:theoretical_verification} (right) also converges to the minimum value in this case, which is proven in \Cref{append:theory}. As such, in the memorization regime when the model has a much larger capacity than the training data, the reproducibility among different diffusion models and the theoretical mapping implies that all diffusion models are approximating the same score function of the empirical multi-delta distribution of the training data. In this regime, the diffusion model lacks the ability to generate new samples.




\subsection{Reproducibility in Generalization Regime} \label{sec:analysis_generalization}

Second, we study reproducibility in the generalization regime, which is the typical training setting for most practical diffusion models. Within this regime, we first focus on examining the learning of score function through model reproducibility. Based upon preliminary studies using simple models, we show that in the generalization regime,
\begin{tcolorbox}
\begin{center}
   \emph{reproducibility is a byproduct of diffusion model learning the \textbf{ground-truth distribution} $p(\bm x_0)$.}
\end{center}
\end{tcolorbox}
Following this, we conduct a thorough investigation into the reproducibility of various pre-trained diffusion models used in real-world applications.



\subsubsection{Reproducibility \& Distribution Learning} \label{sec:analysis_generalization_gt}

However, analysis of the estimation accuracy under the true natural image distribution is exceedingly challenging. Instead, we illustrate through empirical evidence that diffusion models have the capacity to learn the underlying distribution by utilizing data samples generated from two \emph{given} distributions: (\emph{i}) a mixture of Gaussian distribution and (\emph{ii}) pre-trained diffusion models.


\paragraph{Case 1: Learning score functions of a mixture of Gaussians.} We first consider learning diffusion models based upon the following \emph{mixture of low-rank Gaussian} (MoG) distribution:\footnote{As shown in \cite{wang2023hidden}, the learned real data distribution could be approximated as the Mixture of Gaussian distribution.}
\begin{align}\label{eqn:mlg}
    p(\bm x_0) =\frac{1}{C}\sum_{i \in [C]} \mathcal{N}\left(\bm x_0; \bm 0, \mathbf \Sigma_i \right) \;\text{with}\;  \mathbf \Sigma_i = \bm U_i\bm U_i^\top,
\end{align}
where $C$ is the number of classes, and $\bm U_i^* \in \mathbb{R}^{d \times r}$ is the low-rank basis for the $i$th class with $r \ll d$. In this case, by invoking \Cref{lem:key}, we can show that the corresponding score function has the following form.




\begin{proposition}\label{proposition:MoG distribution}

Under the same setting of \Cref{lem:key} with $p(\bm x_0)$ following the MoG distribution introduced in \eqref{eqn:mlg},  we can show that the optimal score function is:
    \begin{align*}
            &\bm s_{\mathrm{MoG}}(\bm x_t, t) = \sum_{i \in [C]} \frac{\pi_i (\bm x_t, t)}{s_t^2\sigma_t^2} \left(- \bm x_t + \frac{1}{1 + \sigma_t^2} \bm U_i\bm U_i^\top \bm x_t\right),
    \end{align*}
    with  $\pi_i (\bm x_t, t) = \dfrac{\mathcal{N}\left(\bm x_t; \bm 0, s_t^2 \bm U_i \bm U_i^\top + s_t^2 \sigma_t^2 \bm I_d \right)}{\sum_{i \in [C]} \mathcal{N}\left(\bm x_t; \bm 0, s_t^2 \bm U_i\bm U_i^\top + s_t^2 \sigma_t^2 \bm I_d\right)}$.
\end{proposition}
\begin{figure}[t]
    \begin{center}
    \includegraphics[width=0.5\linewidth]{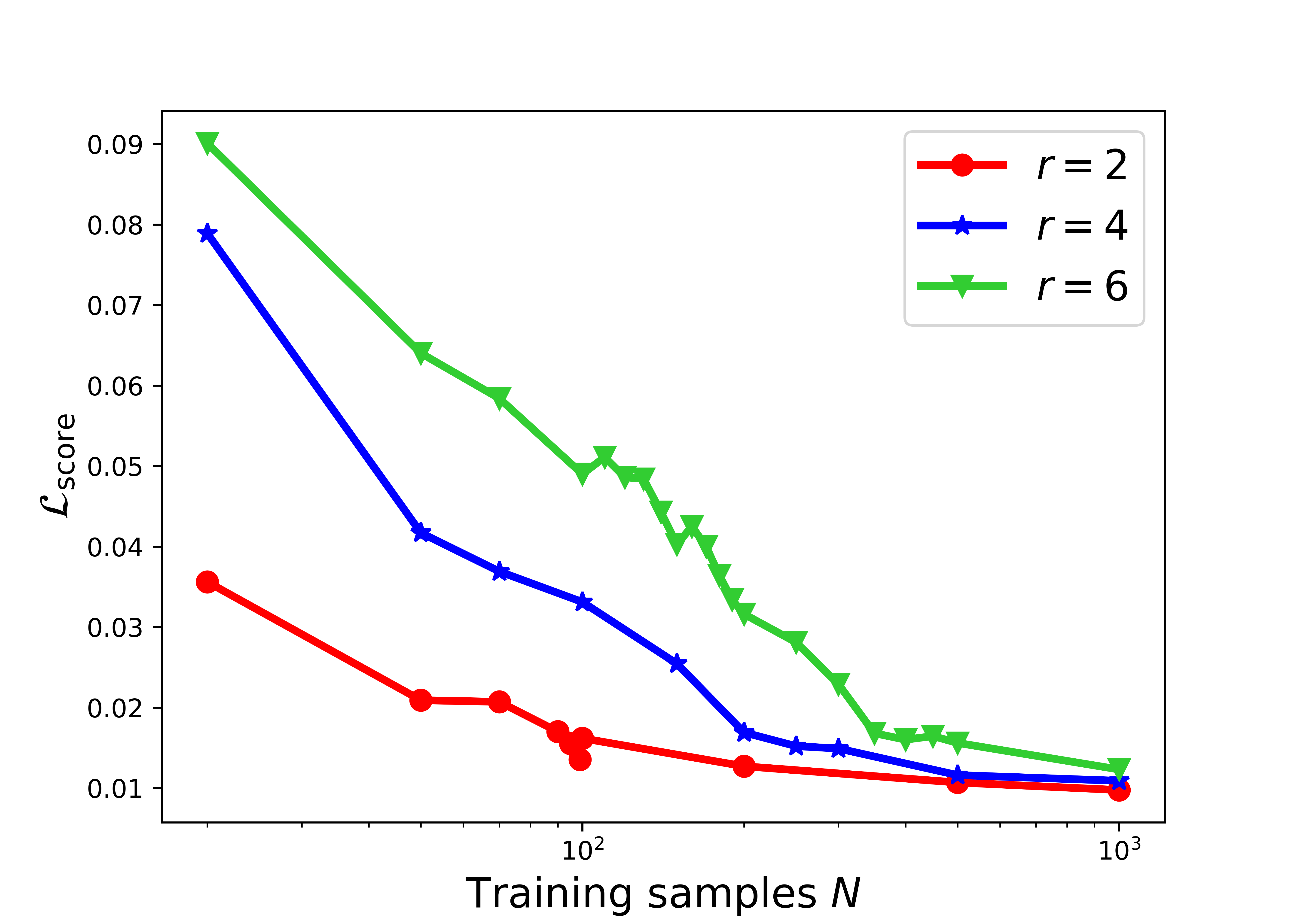}
    \end{center}
    \caption{\textbf{Score matching accuracy.} We train the same diffusion model with varying numbers of training samples $N$ and subspace dimension $r$ from the Mixture of Gaussian distribution defined in \Cref{eqn:mlg} and plot the metric $\mathcal L_{\text{score}}$ in different colors for each $r$. The detailed experimental settings are in \Cref{append:MoG}.}
    \label{fig:toy_model}
\end{figure}
The proof can be found in \Cref{append:theory}.
To test whether practical diffusion models converge to the optimal score function $\bm s_{\mathrm{MoG}}(\bm x_t, t)$, we train the diffusion models $\bm s_{\bm \theta}$ by using $N$ data points $\{\bm y_i\}_{i=1}^N \subseteq \R^n$ drawn from the MoG distribution in \eqref{eqn:mlg}.
We measure the distance between $\bm s_{\mathrm{MoG}}(\bm x_t, t)$ and $\bm s_{\bm \theta}$ by 
\begin{align*}
    &\mathcal L_{\text{score}} := \mathbb{E}_{\substack{t\sim \mathcal U(0, 1), \bm x_0 \sim p(\bm x_0), \bm x_t \sim p_t(\bm x_t |\bm x_0)}} \big[\|\bm s_{\bm \theta}(\bm x_t, t) - \bm s_{\mathrm{MoG}}(\bm x_t, t)\|_2\big],
\end{align*}
where the expectation is calculated for $t$ uniformly sampled from $[0, 1]$, $\bm x_0$ sampled from the MoG distribution $p(\bm x_0)$ and $x_t$ sampled from the noise perturbation kernel $p_t(\bm x_t | \bm x_0)$ given $t$ and $\bm x_0$. From experiment results shown in \Cref{fig:toy_model}, we observe that $\bm s_{\bm \theta}(\bm x_t, t)$ converges to $\bm s_{\mathrm{MoG}}(\bm x_t, t)$ as $N$ increases given different $r$. Therefore, under this setting of MoG distribution, the diffusion model could converge towards the score function $\bm s_{\mathrm{MoG}}$ given enough training samples (in the generalization regime).


\paragraph{Case 2: Learning score functions from pre-trained diffusion models.}
Second, suppose the underlying image distribution $p(\bm x_0)$ can be characterized by the noise-to-image mapping $f_{\bm s_{\bm \theta_1}}(\bm \epsilon), \bm \epsilon \sim \mathcal N(\bm 0, s^2_t \sigma^2_t \bm I_d)$ of a pretrained diffusion model in generalization regime $\bm s_{\bm \theta_1}$. We sample $N$ data points from $p(\bm x_0)$ to generate a training dataset $\{\bm y_i\}_{i=1}^N$, based upon which we train another diffusion model $\bm s_{\bm \theta_2}$ with sufficient large $N$ (in the generalization regime ). We then calculate the reproducibility of the two models following the same metric as in \Cref{sec:metric}.

Experimentally, we find that the two models have a \textbf{high \text{RP Score }$=0.80$}, which indicates that the diffusion model $f_{\bm s_{\bm \theta_2}}$ could converge to the underlying distribution, which is the same data distribution as $f_{\bm s_{\bm \theta_1}}$, and at the same time they have the same noise-to-image mapping. The detailed experiment settings are in \Cref{append:model_recovery}.



\subsubsection{Prevalence of Reproducibility} \label{sec:analysis_generalization_unique}



\begin{figure}[t]
     \centering
     \includegraphics[width=\columnwidth]{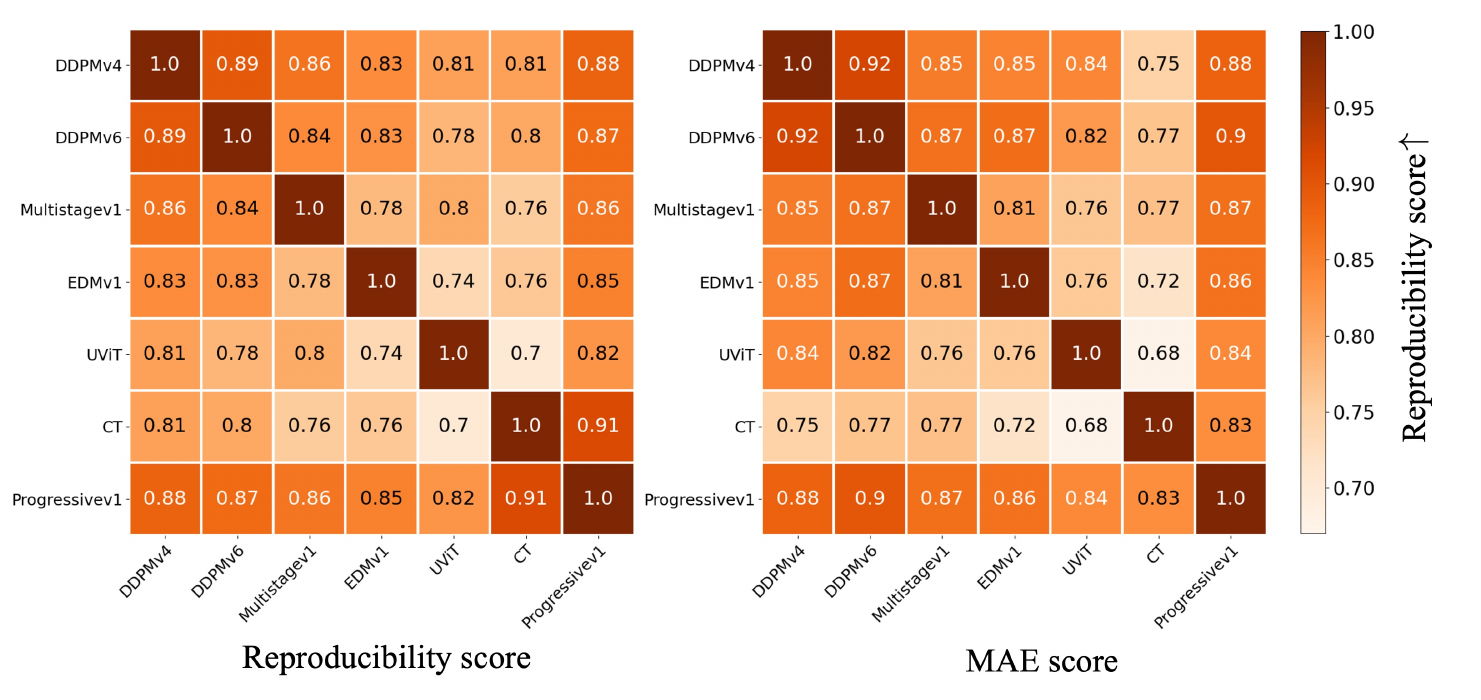}
     \caption{\textbf{Similarity among different unconditional diffusion model settings in generalization regime.} We visualize the quantitative results based upon seven different unconditional diffusion models (DDPMv4, DDPMv6 \cite{ho2020denoising, song2020denoising}, Multistagev1 \cite{multistage}, EDMv1 \cite{karras2022elucidating}, UViT \cite{bao2023all}, CT \cite{song2023consistency}, Progressivev1 \cite{salimans2022progressive}) based upon reproducibility score (left) and MAE score (right) (defined in \Cref{sec:metric}). About more detailed settings and a more comprehensive comparison could be found in \Cref{append:unconditional}.}
     \label{fig:reproducibility_selected}
\end{figure}

Finally, we conclude this section by showing the prevalence of reproducibility in the generalization regime, which is irrespective of \emph{network architectures, training and sampling procedures}, and \emph{perturbation kernels}. Specifically, in \Cref{fig:reproducibility_selected}, we visualize the \emph{similarity matrix} for seven different popular diffusion models, where each element of the matrix measures pairwise similarities of two different diffusion models based upon RP score (left) and MAE score (right). 
All the models are trained with the CIFAR-10 dataset \cite{krizhevsky2009learning}. 
Experimental details and more comprehensive studies can be found in \Cref{append:unconditional}.

As we can see from \Cref{fig:reproducibility_selected}, there is a very consistent model reproducible phenomenon for comparing any two models. For even the most dissimilar models, the RP and MAE scores are notably high at 0.7 and 0.68, respectively. Specifically, we observe the following:
\begin{itemize}[leftmargin=*]
    \item \textbf{Different network architectures.} We evaluate (i) U-Net \cite{ronneberger2015u} based architecture: DDPM \cite{ho2020denoising}, DDPM++ \cite{song2020score}, Multistage \cite{multistage}, EDM \cite{karras2022elucidating}, Consistency Training (CT) and Distillation (CD) \cite{song2023consistency}, and (ii) Transformer \cite{vaswani2017attention} based architecture: DiT \cite{peebles2022scalable} and U-ViT \cite{bao2023all}. This phenomenon remains consistent regardless of the specific architecture employed.   
    \item \textbf{Different training procedures.} We consider discrete \cite{ho2020denoising} and continuous \cite{song2020score} settings, training from scratch or distillation \cite{salimans2022progressive, song2023consistency} for the diffusion model.  When we compare CT (consistency training) and EDMv1, even when we use different training losses, they both converge to similar noise-to-image mappings.  Additionally, comparing DDPMv1 and Progressivev1 reveals that both training from scratch and distillation approaches lead to the same results.
    
    \item \textbf{Different sampling procedures.} For sampling, we only use \emph{deterministic} samplers, such as DPM-Solver \cite{lu2022dpm}, Heun-Solver \cite{karras2022elucidating}, DDIM \cite{song2020denoising} etc. For example, DDPMv4 utilizes DPM-solver, EDMv1 employs a 2nd order heun-solver, and CT utilizes consistency sampling, yet they all exhibit very high model reproducibility.
    \item \textbf{Different perturbation kernels.} For the data corruption process, we compared Variance Preserving (VP) \cite{ho2020denoising}, Variance Exploding (VE), and sub Variance Preserving (sub-VP) \cite{song2020score} perturbation methods for noise perturbation stochastic differential equations. We scale the initial noise using the standard deviation specific to the terminated Gaussian distribution of each perturbation kernel to ensure a fair comparison, details can be found in \Cref{append:unconditional}. Our observations indicate that the choice of perturbation methods (VP, sub-VP, and VE) has a limited impact on reproducibility when comparing DDPMv4, DDPMv6, and EDMv1.

\end{itemize}

\subsubsection{Reproducibility from Noise Hyperplane to Image Manifold}

\begin{figure}[t]
     \centering
     \centering
     \includegraphics[width=1.0\textwidth]{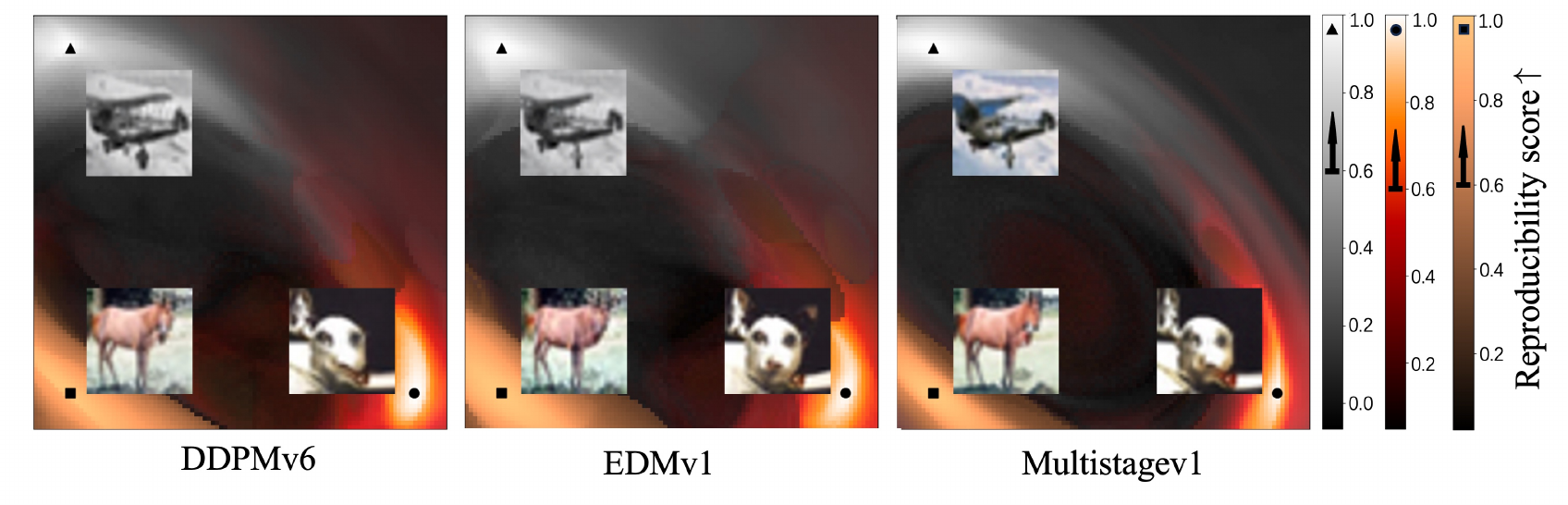}
     \caption{\textbf{Unqiue Encoding from Noise Hyperplane to Image Manifold.} The diagram illustrates the process of mapping from a noise hyperplane to the image manifold. We employ three distinct models: DDPMv6, EDMv1, and Multistagev1. Initially, we select three different initial noises from Gaussion and generate corresponding samples, denoted by a triangle, square, and circle in the first three images on the left. The hyperplane is defined based on these chosen noises. Image generations, starting from uniformly selected initial noise within this hyperplane, are classified as identical to either the triangle, square, or circle image, determined by the maximum SSCD similarity with them. Each initial noise is colored according to its generation's corresponding class (as indicated on the right; for instance, the noise's generation identical to the triangle image is represented by the black-white color bar), along with the SSCD similarity to the identical image. 
     }
     \label{fig:MappingManifold}
\end{figure}

While \Cref{sec:analysis_generalization_unique} studies the reproducibility of images generated from discretized initial noises $\bm \epsilon \in \mathcal E$. In this section, we want to further explore the reproducibility of the image manifold generated from 2D noise hyperplane $\mathcal H \subseteq \mathcal E$. Specifically, we find that  
\begin{itemize}[leftmargin=*]
    \item \emph{Similar unique encoding maps across different network architectures.} We further confirm the model reproducibility by visualizing the generated image manifold from the same 2D noise hyperplane, inspired by \cite{somepalli2022can}. The visualization in \Cref{fig:MappingManifold} shows that different generated manifolds of different network architectures share very similar structures.
    
    \item \emph{Local Lipschitzness of the unique encoding from noise to image space.} Furthermore, our visualization suggests that $f_{\bm s}$ is locally Lipschitz, where $\| f_{\bm s}(\bm \eps_1) - f_{\bm s}(\bm \eps_2) \| \leq L \| \bm \eps_1 - \bm \eps_2 \|$ for any $\bm \eps_1,\bm \eps_2 \in \mathcal B(\bm \eps, \delta_{\bm \eps}) \cap \mathcal E$ with some Lipschitz constant $L$. Here $\mathcal B(\bm \eps, \delta_{\bm \eps})$ denotes a ball centered at a Gaussian noise $\bm \eps$ with radius $\delta_{\bm \eps}$. In other words, noises $\bm \eps_1,\bm \eps_2 \in \mathcal E$ close in distance would generate similar reproducible images in $\mathcal I$ via diffusion models.
\end{itemize}

Specifically, the visualization in \Cref{fig:MappingManifold} is created as follows. First, we pick three initial noises $(\bm{\epsilon}_1, \bm{\epsilon}_2, \bm{\epsilon}_3 ) $ in the noise space $\mathcal E$ and used different diffusion model architectures to generate clear images ($\bm x_1$, $\bm x_2$, $\bm x_3$) in the image manifold $\mathcal I$, so that the images $\{\bm x_i\}_{i=1}^3$ belong to three different classes. Second, we create a 2D noise hyperplane with
\begin{align*}
   \bm{\epsilon}\paren{\alpha, \beta} = \alpha \cdot (\bm{\epsilon}_2 - \bm{\epsilon}_1) + \beta \cdot (\bm{\epsilon}_3 - \bm{\epsilon}_1) + \bm{\epsilon}_1
\end{align*}
Within the region $ (\alpha,\beta) \in [-0.1,1.1] \times [-0.1,1,1] $, we uniformly sample $100 $ points along each axis and generate images  $\bm x\paren{\alpha, \beta}$ for each sample $\bm{\epsilon}\paren{\alpha, \beta}$ using different diffusion model architectures (i.e., DDPMv6, EDMv1, Multistagev1). For each point $\paren{\alpha, \beta}$, it is considered as identical to image $\bm x_i$ for $i=\argmax_{k\in\{1, 2, 3\}} [\mathcal{M}_{\text{SSCD}}(\bm x_k, \bm x\paren{\alpha, \beta})]$, and we visualize the value of $\mathcal{M}_{\text{SSCD}}(\bm x_i, \bm x\paren{\alpha, \beta})$. As we observe from \Cref{fig:MappingManifold}, the visualization shares very similar structures across different network architectures. Second, for each plot, closeby noises create images with very high similarities. These observations support our above claims.

%% file: section_new/4.phenomenon_extension.tex
\section{Beyond Unconditional Diffusion Models}
\label{sec:reproducibility_more}


\begin{figure*}[t]
     \centering
     \includegraphics[width=\linewidth]{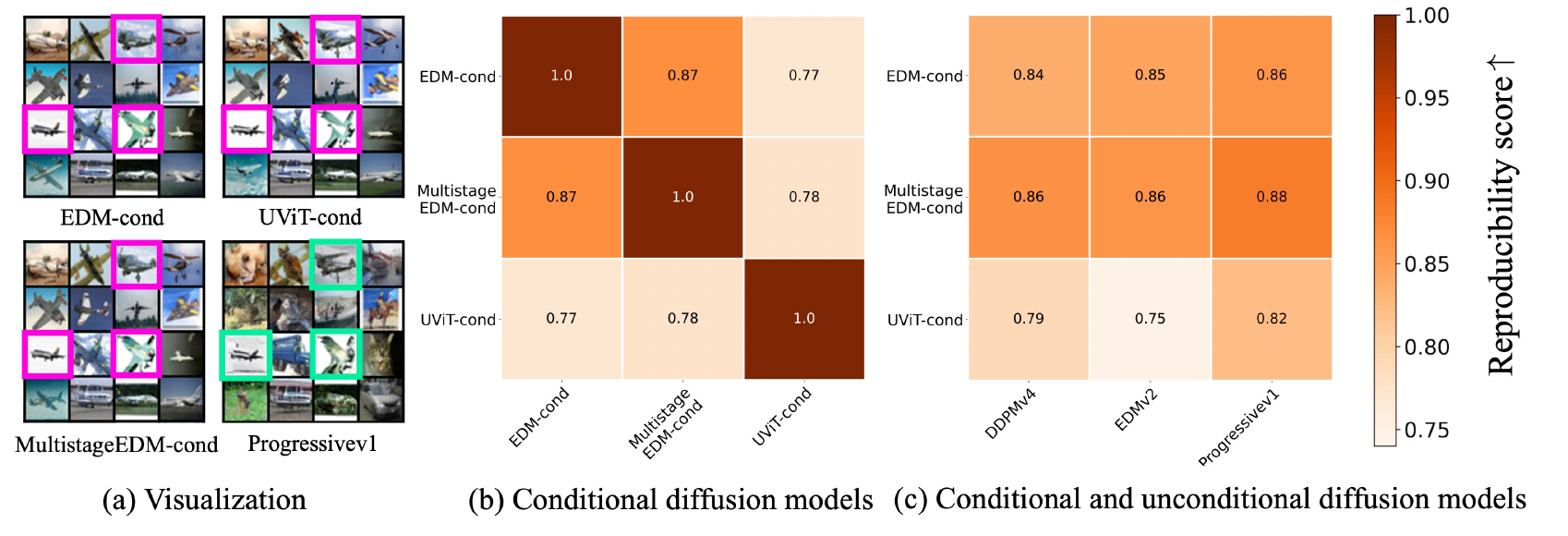}
     \caption{\textbf{Model reproducibility for conditional diffusion model in the generalization regime. } In this study, we employ conditional diffusion models, specifically U-Net-based (EDM-cond, MultistageEDM-cond) and transformer based (UViT-cond), which we train on the CIFAR-10 dataset using class labels as conditions.  Additionally, we select unconditional diffusion models, namely Progressivev1, DDPMv4, and EDMv2, as introduced in  \Cref{sec:analysis_generalization_unique}. Figure (a) showcases sample generations from both unconditional and conditional diffusion models (with the "plane" serving as the condition for the latter). Notably, samples within the same row and column originate from the same initial noise. The reproducibility scores between the conditional diffusion models are presented in (b), and between unconditional and conditional diffusion models in (c).}
     \label{fig:rp_conditional}
\end{figure*}

In this section, we explore the concept of model reproducibility in a broader context, extending beyond unconditional diffusion models. We demonstrate that model reproducibility manifests \textbf{more generally} across various scenarios, including conditional diffusion models, diffusion models for inverse problems, and the fine-tuning of diffusion models.



\begin{figure}[t]
     \centering
     \begin{subfigure}[t]{0.6\columnwidth}
         \centering
         \includegraphics[width=\textwidth]{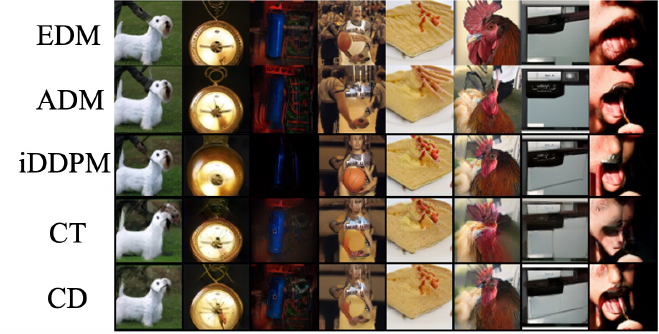}
         \caption{Visualization.}
         \label{fig:imagenet_visadm}
     \end{subfigure}
     \hfill 
     \begin{subfigure}[t]{0.38\linewidth}
         \centering
         \includegraphics[width=\linewidth]{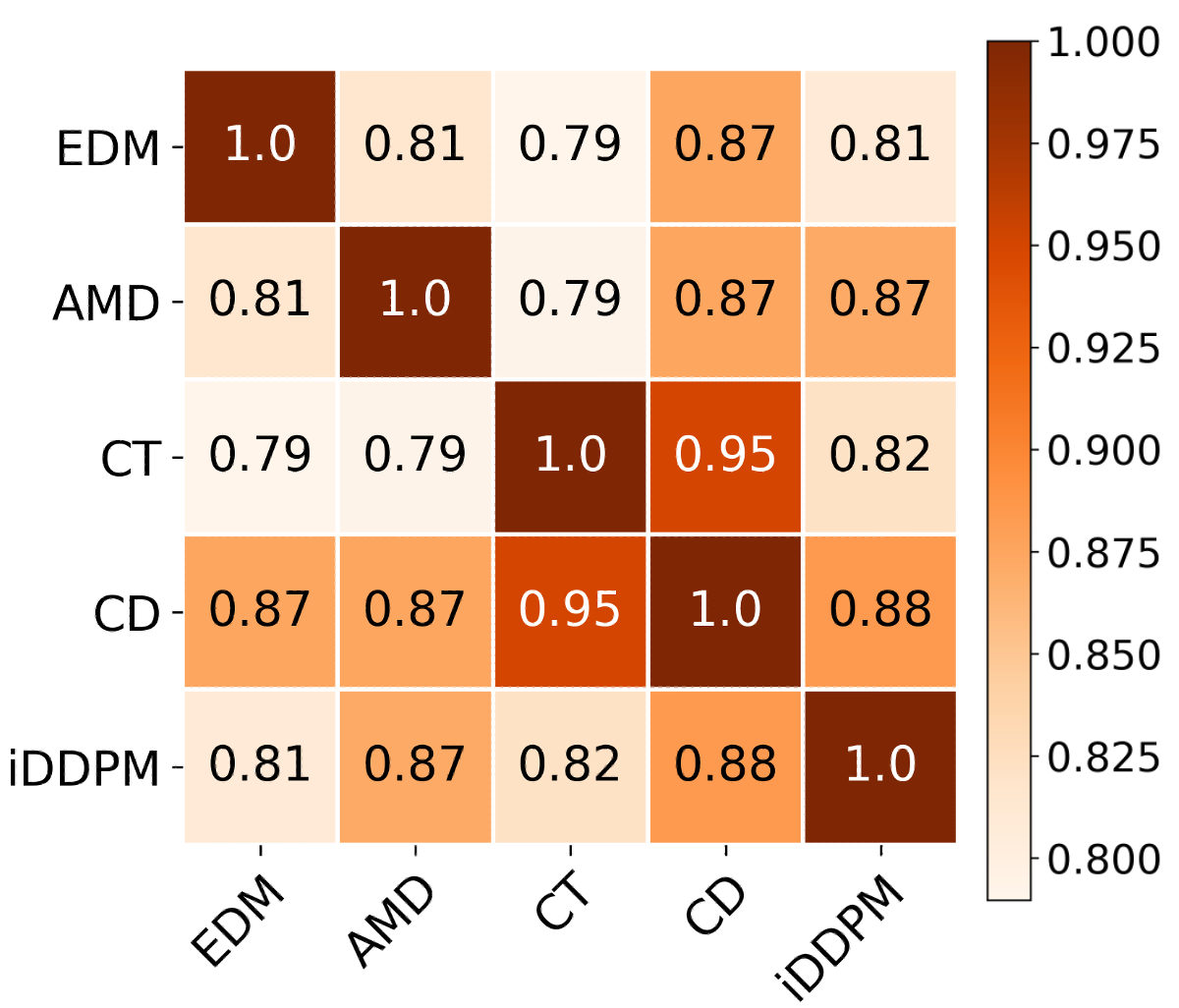}
         \caption{$\text{RP}_{cond}$ Score}
         \label{fig:imagenet_reproducibility}
     \end{subfigure}
     \caption{\textbf{Model reproducibility for conditional diffusion model generations on ImageNet dataset.} In this experiment, we choose the conditional diffusion model (EDM, ADM \cite{dhariwal2021diffusion}, CD, CT, iDDPM \cite{nichol2021improved}) trained on the ImageNet dataset. 10K image pairs are generated to estimate the $\text{RP}_{cond}$ Score. Due to the complexity of the ImageNet dataset, we set the threshold for the SSCD metric as 0.4 instead of 0.6 here, following the setting in \cite{somepalli2023understanding}.}
     \label{fig:imagenet}
\end{figure}

\begin{figure*}[t]
     \centering
     \includegraphics[width=\linewidth]{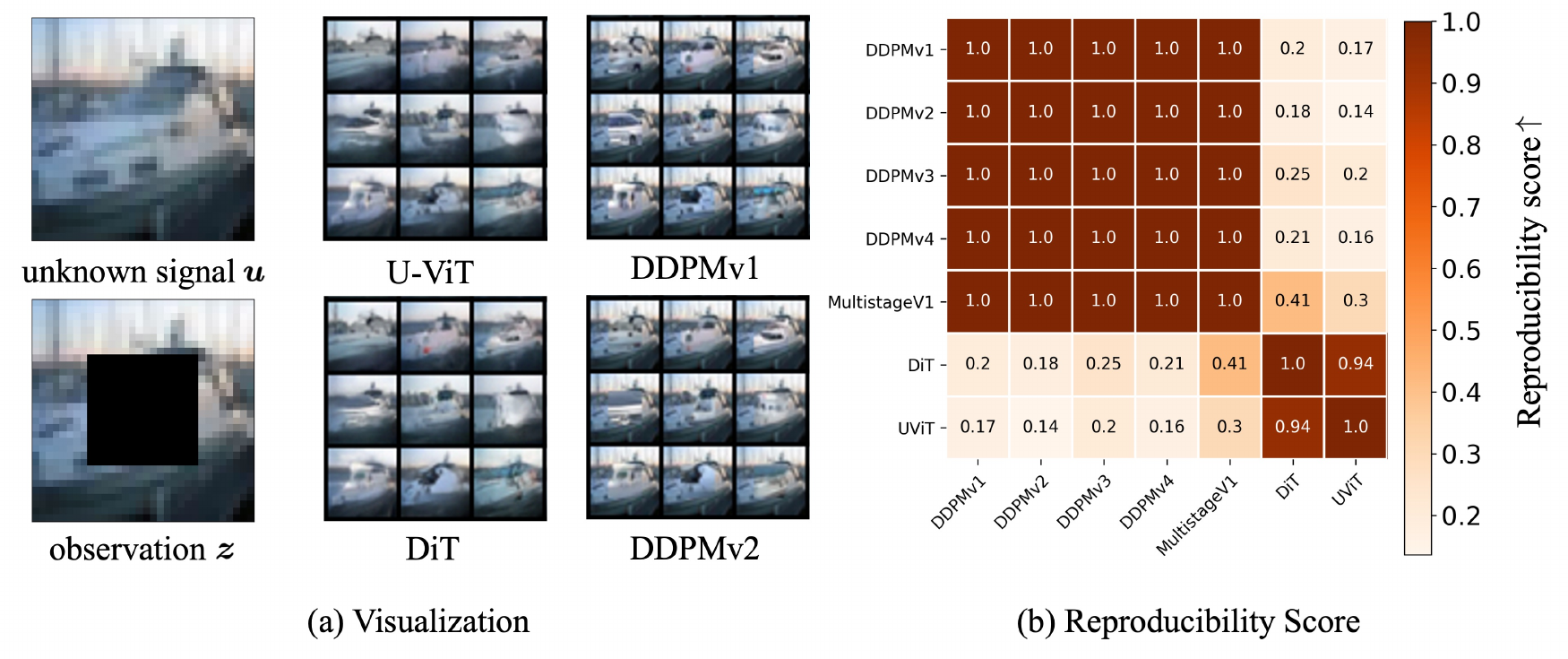}
     \caption{\textbf{Model reproducibility for solving inverse problems in the generalization regime.} In this investigation, we employ various unconditional diffusion models, as introduced in \Cref{sec:analysis_generalization_unique}, which were initially trained on the CIFAR-10 dataset. Our approach involves utilizing a modified deterministic variant of diffusion posterior sampling (DPS), as detailed in \Cref{append:inverseproblem}. Specifically, we focus on the task of image inpainting. Figure (a) presents both the observation $\bm z$, unknown signal $\bm u$, and generations from different diffusion models. Notably, samples within the same row and column originate from the same initial noise. The reproducibility scores for different diffusion models under the DPS algorithm are quantitatively analyzed in (b).}
     \label{fig:rp_inverseproblem}
\end{figure*}

\subsection{Conditional Diffusion Models} 
Conditional diffusion model, introduced by \cite{ho2022classifier, dhariwal2021diffusion}, gained its popularity in many applications such as text-to-image generation \cite{rombach2022high, ramesh2021zero, nichol2021glide}. These models achieve a superior degree of control and enhanced quality in output generation through the integration of rich class embeddings within the denoising function. Interestingly, we find that:

\begin{tcolorbox}
    \begin{center}
    \emph{Model reproducibility of conditional models exhibits in a structured way and is strongly related to unconditional counterparts.}
    \end{center}
\end{tcolorbox}

Specifically, our experiments in \Cref{fig:rp_conditional} demonstrate that (\emph{i}) model reproducibility exists among different conditional diffusion models, and (\emph{ii}) model reproducibility presents between conditional and unconditional diffusion models \emph{only} if the type (or class) of content generated by the unconditional models matches that of the conditional models. More results can be found in  \Cref{append:conditional}.

To support our claims, we define the \emph{conditional reproducibility score} between different conditional diffusion models by $\text{RP}_{cond} \text{ Score} := \mathbb{P}\paren{\mathcal{M}_{\text{SSCD}}(\bm x^c_1, \bm x^c_2)>0.6 \mid c \in \mathcal C }$ to evaluate similarity between outputs of different conditional diffusion models, based on the likelihood of their similarity exceeding a threshold from the same initial noise and conditioned on the class $c\in \mathcal C$. We also introduce a between reproducibility score $\text{RP}_{between} \text{ Score}:= \mathbb{P}\paren{\max_{_{c \in \mathcal C}}[\mathcal{M}_{\text{SSCD}}(\bm x_1, \bm x^c_2)]>0.6}$, for an unconditional generation $\bm x_1$ and conditional generation $\bm x^c_2$ originating from an identical noise, to assess the similarity between unconditional output $\bm x_1$ and conditional output $\bm x^c_2$. 


Results in \Cref{fig:rp_conditional} (a) (b) show that samples from different conditional models (EDM-cond, UViT-cond, MultistageEDM-cond) are similar when conditioned on the same class and noise, supporting Claim (i). On the other hand, a high $\text{RP}_{between} \text{ Score}$ and visual similarities between unconditional and conditional samples, as seen \Cref{fig:rp_conditional} (c), support Claim (ii). Furthermore, beside the CIFAR-10 dataset, we also demonstrate the conditional reproducibility on large-scale datasets such as ImageNet \cite{deng2009imagenet} in \Cref{fig:imagenet} and large-scale diffusion models such as Stable Diffusion \cite{rombach2022high} in \Cref{append:stablediffusion}.


\subsection{Diffusion Models for Solving Inverse Problems}
Recently, diffusion models have also been demonstrated as rich, structural priors to solve a broad spectrum of inverse problems \cite{song2023solving, chung2022diffusion, song2021solving, chung2022improving},\footnote{Here, the problem is often to reconstruct an unknown signal $\bm u$ from the measurements $\bm z$ of the form $\bm z = \mathcal{A}(\bm u) + \bm \eta$, where $\mathcal{A}$ denotes some (given) sensing operator and $\bm \eta$ is the noise.} including but not limited to image super-resolution, de-blurring, and inpainting. Motivated by these promising results, our illustration is based upon solving the image inpainting problem using a modified deterministic variant of diffusion posterior sampling (DPS) \cite{chung2022diffusion}, showcasing that for solving inverse problems using diffusion models: 

\begin{tcolorbox}
\begin{center}
    \emph{Model reproducibility holds only within the same type of network architectures.} 
\end{center}
\end{tcolorbox}

Our claim is supported by the experimental results in \Cref{fig:rp_inverseproblem}.
Specifically, \Cref{fig:rp_inverseproblem} (a) virtualizes the samples generated from different diffusion models, and \Cref{fig:rp_inverseproblem} (b) presents the similarity matrix of model reproducibility between different models, i.e., U-Net based (DDPMv1, DDPMv2, DDPMv3, DDPMv4, Multistagev1) and Transformer based (DiT, U-ViT) architectures. We note a strong degree of model reproducibility \emph{among} architectures of the same type (e.g., U-Net vs. Transformer), but the model reproducibility score exhibits a notable decrease when any U-Net model is compared with any Transformer-based model.

We conjecture that the lack of reproducibility across network architectures is due to the following reasons: (\emph{i}) DPS introduces the gradient term $\frac{\partial \bm s_{\bm \theta}(\bm x_t,t)}{\partial \bm x_t}$ during the sampling, and this extra term might break the reproducibility for different type of architectures. (\emph{ii}) the reproducibility between different types of architectures might not hold for out-of-distribution data generation, whereas the data $\bm x_t$ passed into the learned score function $\bm s_{\bm \theta}(\bm x_t,t)$ is out-of-distribution for solving inverse problems. We leave these for future study.

\subsection{Model Reproducibility in Fine-tuning Diffusion Models.}

\begin{figure*}[t]
     \centering
     \begin{subfigure}[t]{0.9\textwidth}
         \centering
         \includegraphics[width=\linewidth]{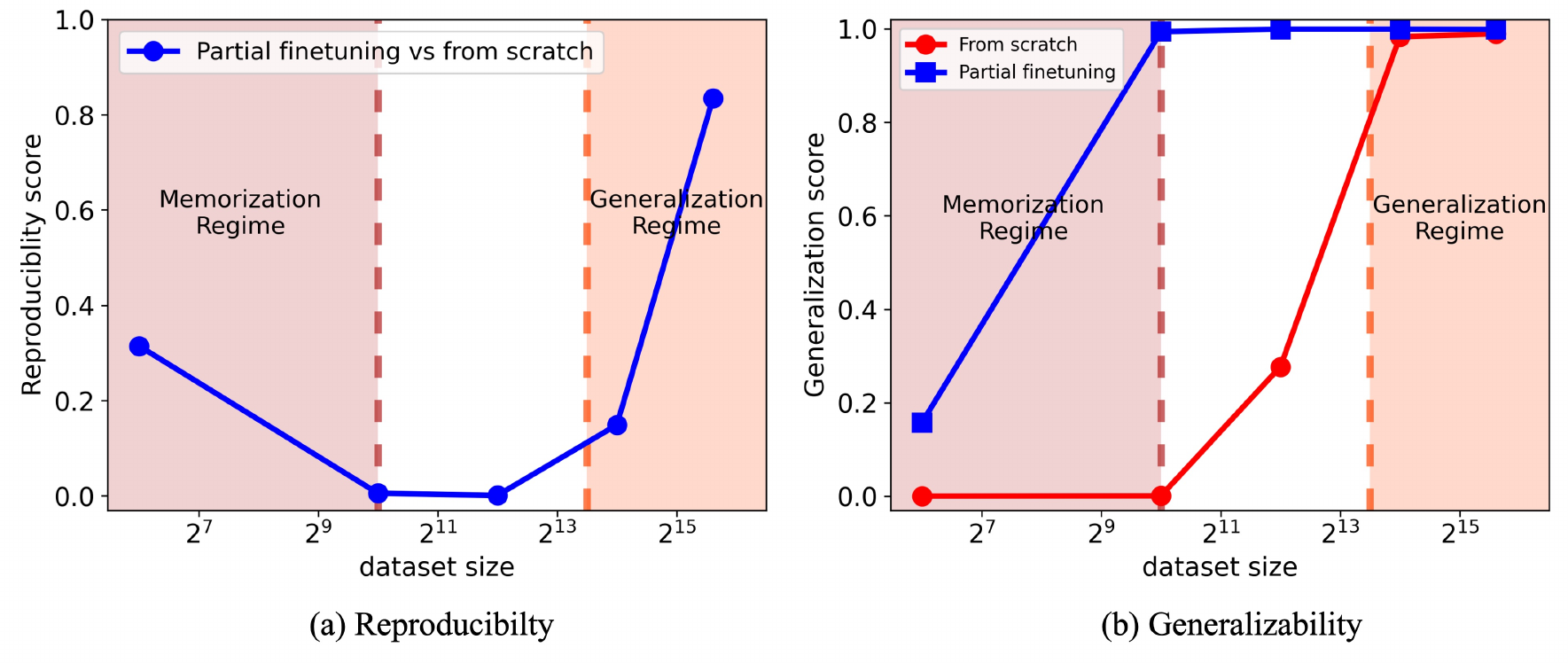}
         \label{fig:reproducibility_finetuning}
     \end{subfigure}
     \caption{\textbf{Model reproducibility for diffusion model finetuing.} In this experiment, we employ DDPMv4. Two distinct training strategies are investigated: "from scratch," denoting direct training on a subset of the CIFAR-10 dataset, and "partial fine-tuning," which involves pretraining on the entire CIFAR-100 dataset \cite{krizhevsky2009learning} followed by fine-tuning only the attention layers of the model on a subset of the CIFAR-10 dataset. The dataset sizes for CIFAR-10 range from $2^6$ to $2^{15}$. Importantly, both "from scratch" and "partial fine-tuning" are trained using the same subset of images for each dataset size. Under different dataset sieze, Figure (a) illustrates the reproducibility score between these two strategies and (b) presents the generalization score for them.}    
     \label{fig:rp_finetuning}
\end{figure*}

Few-shot image fine-tuning for diffusion models, as discussed in \cite{ruiz2023dreambooth, gal2022image, moon2022finetuning, han2023svdiff}, showcases remarkable generalizability. This is often achieved by fine-tuning a small portion of the parameters of a large-scale pre-trained (text-to-image) diffusion model. In this final study, we delve into the impacts of partial model fine-tuning on both model reproducibility and generalizability, by extending our analysis in \Cref{sec:tworegime}. We show that:
\begin{tcolorbox}
    \begin{center}
    \emph{Partial fine-tuning reduces reproducibility but improves generalizability in ``memorization regime''.}
    \end{center}
\end{tcolorbox}

Our claim is supported our results in \Cref{fig:rp_finetuning}, comparing model fine-tuning and training from scratch of with varying size of the training data, where both models have the same number of parameters. In comparison to training from scratch that we studied in \Cref{fig:regime}, fine-tuning specific components of pre-trained diffusion models, particularly the attention layer in the U-Net architecture, yields lower model reproducibility score but higher generalization score in the memorization regime. However, in the generalization regime, partial model fine-tuning has a minor impact on both reproducibility and generalization in the diffusion model.
Our result reconfirms the improved generalizability of fine-tuning diffusion models on limited data, but shows a surprising tradeoff in terms of model reproducibility that is worth of further investigations.


%% file: section_new/5.related_works.tex
\section{Related Works}\label{sec:related}

\paragraph{Convergence analysis of diffusion models.}
Numerous theoretical studies have investigated the diffusion model's convergence towards the underlying distribution. Most of these studies, including \cite{wellscore1, wellscore2, wellscore3, wellscore4, wellscore5, wellscore6}, have established convergence by assuming an $L^2$-accurate score estimation. Others have explored convergence without relying on this assumption. Nonetheless, these studies rely on strong simplification regarding network architectures \cite{li2023generalization, chen2023score} and data distributions \cite{chen2023score}. Our paper provides an empirical complement to existing theoretical analyses. 

In contrast, our paper focuses on the learned distribution and score function under various practical diffusion model settings. The empirical findings not only broaden the understanding of diffusion models in realistic settings but also bridge the gap between theory and practice.

\paragraph{Understanding memorization \& generalization.}

Recently, \cite{yoon2023diffusion} categorized the training regimes of diffusion models into memorization and generalization, concluding that diffusion models tend to generalize when they fail to memorize the training data. In the memorization regime, \cite{yi2023generalization, gu2023memorization} demonstrated that training diffusion models converges towards an optimal denoiser. In contrast, in the generalization regime, \cite{pidstrigach2022score} linked generalization in simple settings to avoiding overfitting, while \cite{kadkhodaie2023generalization} showed that the generalization capabilities of diffusion models arise from an implicit bias towards geometry-adaptive harmonic bases. Furthermore, \cite{somepalli2023diffusion, somepalli2023understanding, carlini2023extracting} revealed that diffusion models can still replicate training samples even in the generalization regime, leading to significant privacy concerns.

In comparison, our work takes a step further to delve into the problem. By examining the largely overlooked reproducibility phenomenon, our work is the first to show that diffusion models learn distinct distributions in different regimes: in the memorization regime, diffusion models learn the empirical distribution, while in the generalization regime, they learn the underlying distribution. Moreover, our research provides the first empirical evidence that diffusion models can overcome the curse of dimensionality when learning the underlying distribution, enabling effective generalization even with a limited number of training samples. Finally, our analysis also extends to conditional diffusion models and diffusion models for inverse problems, which have not been addressed in previous studies.

\paragraph{Reproducibility in deep learning.}
Theoretically, the reproducibility we identified for diffusion models is similar to the notion of unique identifiable encoding, that several prior studies have explored for deep latent-variable models. This property refers to the ability of models to converge to a specific input-embedding mapping, irrespective of variations in weight initialization or optimization methods \cite{roeder2021linear}. The foundational work for this property in deep latent-variable models was established through the analysis of Independent Component Analysis (ICA) by \cite{hyvarinen2016unsupervised, hyvarinen2017nonlinear, hyvarinen2019nonlinear}. Building upon this, \cite{khemakhem2020variational} demonstrated the identifiability of Variational Autoencoders (VAE) using conditionally factorial priors over latent variables, while \cite{roeder2021linear} provided evidence of linear identifiability in representation learning. Empirically, studies such as \cite{li2015convergent} and \cite{somepalli2022can} have observed reproducibility in representation learning and classification tasks, respectively, using similar network architectures but different training procedures. 

While \cite{song2020score} mentioned that diffusion models possess the property of unique identifiable encoding, our novel empirical findings show that diffusion models consistently converge to a similar noise-to-image mapping. This occurs regardless of variations in network architectures, noise perturbation kernels, or training and sampling procedures.

%% file: section_new/6.discussion.tex
\section{Conclusions and Implications}
\label{sec:conclusion}

This study focuses on an important phenomenon in diffusion models, which we term “consistent model reproducibility”. We believe this intriguing phenomenon could significantly impact future research on diffusion models. Below, we outline several promising directions:

\paragraph{Improving training efficiency.}

The potential of this work to improve the training efficiency of diffusion models lies in leveraging the distinct relationship between noise and image spaces. Recent research \cite{multistage} illustrates this by delineating the training of diffusion models into three stages, each employing networks of varying sizes. This approach capitalizes on the reproducibility phenomenon, indicating that adequately parameterized networks learn the same score function. Consequently, by appropriately adjusting parameter sizes for each stage, empirical evidence shows that the proposed method surpasses existing techniques, particularly in improving training efficiency in the generalization regime. These findings imply that incorporating reproducibility as a guiding principle in training diffusion models holds significant promise for future research endeavors.

\paragraph{Black-box model privacy.} 
Several commercial, large-scale diffusion models, e.g. Imagen \cite{saharia2022photorealistic}, DALL-E \cite{betker2023improving}, are designed as black-box systems, raising significant privacy concerns due to the property of reproducibility. Our analysis, in the Case 2 of \Cref{sec:analysis_generalization_gt}, indicates that one can replicate the mapping from a trained diffusion model $f_{\bm s_{\bm \theta}}$ by training a new score function 
 $\bm s_{\bm \theta'}$ from generated data by $f_{\bm s_{\bm \theta}}$ (through the open-source API). Furthermore, given the white-box duplication $f_{\bm s_{\bm \theta'}}$, gradient-based adversarial attacking \cite{guo2021gradient} and training data privacy \cite{carlini2023extracting} would arise as more exacerbated problems. 
\paragraph{Controllable data generation.}
Given the unique mapping learned by the diffusion model, we could control image distribution by manipulating the noise distribution. More specifically, in text-driven image generation, image distribution could be manipulated for adversarial attacking \cite{zou2023universal}, robust defending \cite{zhu2023promptbench}, copyright protection \cite{somepalli2023understanding, somepalli2023diffusion}. In solving inverse problems, one recent paper \cite{liu2023accelerating} manipulated the noise distribution for more efficient sampling. Beyond, the image distribution could also be designed to reduce the uncertainty and variance in our signal reconstruction ~\cite{langevin,chung2022score-MRI,luo2023bayesian}.

%% file: section_new/appendix/Appendix_unconditional.tex


\vspace{0.2in}

\section{Unconditional Diffusion Model}\label{append:unconditional}

\paragraph{Expanded experiment setting}

More detailed settings of the diffusion model we selected are listed in \Cref{appendtab:unconditional_setting}. With the exception of DiT and UViT, which we implemented and trained ourselves, all selected diffusion model architectures utilize the author-released models.

\paragraph{Architectural Relationships}
For DDPMv1, DDPMv2, and DDPMv7, we adopt the DDPM architecture initially proposed by \cite{ho2020denoising}, but we implement it using the codebase provided by \cite{song2020score}. DDPMv3 and DDPMv8, on the other hand, employ DDPM++, an enhanced version of DDPM introduced by \cite{song2020score}. DDPM++ incorporates BigGAN-style upsampling and downsampling techniques, following the work of \cite{brock2018large}. DDPMv4, DDPMv5, and DDPMv6 adopt DDPM++(deep), which shares similarities with DDPM++ but boasts a greater number of network parameters. Moving to Multistagev1, Multistagev2, and Multistagev3, these models derive from the Multistage architecture, a variant of the U-Net architecture found in DDPM++(deep). For EDMv1, EDMv2, CT, and CD, the EDM architecture is identical to DDPM++, but they differ in their training parameterizations compared to other DDPM++-based architectures. Finally, UViT and DiT are transformer-based architectures.

\paragraph{Distillation Relationships}

CD, Progressivev1, Progressivev2, and Progressivev3 are all diffusion models trained using distillation techniques. CD employs EDM as its teacher model, while Progressivev1, Progressivev2, and Progressivev3 share DDPMv3 as their teacher model. It's worth noting that these models employ a progressive distillation strategy, with slight variations in their respective teacher models, as elaborated in \cite{salimans2022progressive}.

\paragraph{Initial Noise Consistency}

However, it is important to note a nuanced difference related to the noise perturbation kernels. Specifically, for VP and subVP noise perturbation kernels, we define the noise space as $\mathcal E = \mathcal{N}(\bm 0, \bm I)$, whereas the VE noise perturbation kernel introduces a distinct noise space with $\mathcal E = \mathcal{N}(\bm 0, \bm \sigma^2_{\text{max}} \cdot I)$, where $\sigma_{\text{max}}$ is predefined. So during the experiment, we sample 10K initial noise $\bm \epsilon_{\text{vp, subvp}} \sim \mathcal{N}(\bm 0, \bm I)$ for the sample generation of diffusion models with VP and subVP noise perturbation kernel. For diffusion models with VE noise perturbation kernel, the initial noise is scaled as $\bm \epsilon_{\text{ve}} =  \sigma_{\text{max}} \bm \epsilon_{\text{vp, subvp}}$. 

\begin{figure}
     \centering
     \begin{subfigure}[t]{0.20\textwidth}
         \centering
         \includegraphics[width=\textwidth]{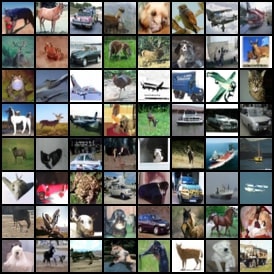}
         \caption{DDPMv1}
     \end{subfigure}
     \hspace{5mm}
     \begin{subfigure}[t]{0.20\linewidth}
         \centering
         \includegraphics[width=\linewidth]{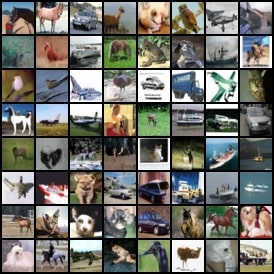}
         \caption{DDPMv2}
     \end{subfigure}
     \hspace{5mm}
     \begin{subfigure}[t]{0.20\linewidth}
         \centering
         \includegraphics[width=\linewidth]{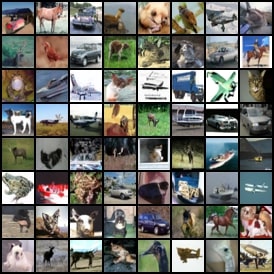}
         \caption{DDPMv3}
     \end{subfigure} 
     \\
     \centering
     \begin{subfigure}[t]{0.20\textwidth}
         \centering
         \includegraphics[width=\textwidth]{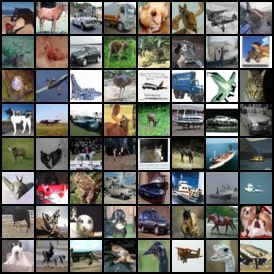}
         \caption{DDPMv5}
     \end{subfigure}
     \hspace{5mm}
     \begin{subfigure}[t]{0.20\linewidth}
         \centering
         \includegraphics[width=\linewidth]{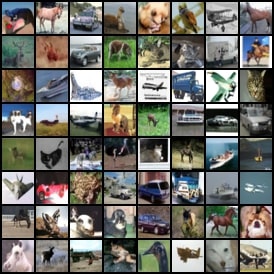}
         \caption{DDPMv6}
     \end{subfigure}
     \hspace{5mm}
     \begin{subfigure}[t]{0.20\linewidth}
         \centering
         \includegraphics[width=\linewidth]{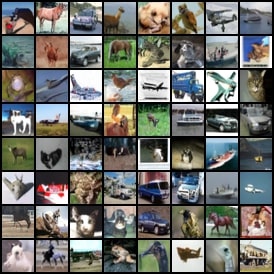}
         \caption{DDPMv7}
     \end{subfigure}     
     \\
     \centering
     \begin{subfigure}[t]{0.20\textwidth}
         \centering
         \includegraphics[width=\textwidth]{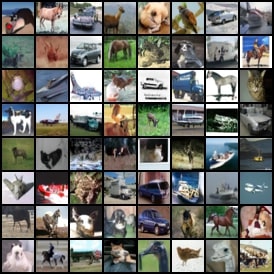}
         \caption{DDPMv8}
     \end{subfigure}
     \hspace{5mm}
     \begin{subfigure}[t]{0.20\linewidth}
         \centering
         \includegraphics[width=\linewidth]{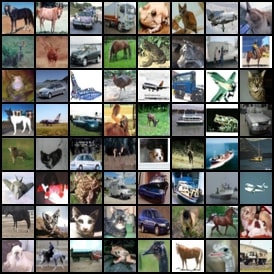}
         \caption{EDMv1}
     \end{subfigure}
     \hspace{5mm}
     \begin{subfigure}[t]{0.20\linewidth}
         \centering
         \includegraphics[width=\linewidth]{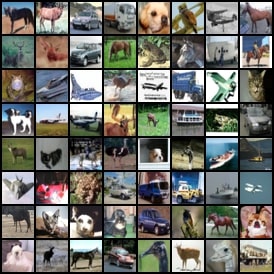}
         \caption{EDMv2}
     \end{subfigure}     
     \\
     \centering
     \begin{subfigure}[t]{0.20\textwidth}
         \centering
         \includegraphics[width=\textwidth]{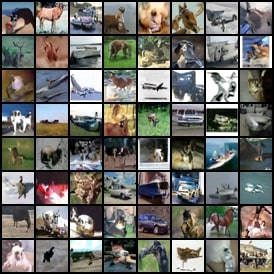}
         \caption{DiT}
     \end{subfigure}
     \hspace{5mm}
     \begin{subfigure}[t]{0.20\linewidth}
         \centering
         \includegraphics[width=\linewidth]{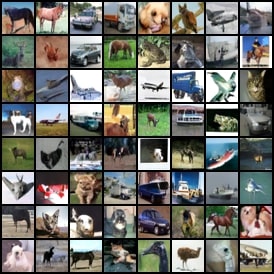}
         \caption{CD}
     \end{subfigure}
     \hspace{5mm}
     \begin{subfigure}[t]{0.20\linewidth}
         \centering
         \includegraphics[width=\linewidth]{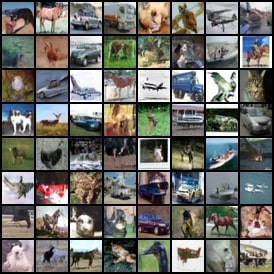}
         \caption{Progressivev1}
     \end{subfigure}     
     \\
     \centering
     \begin{subfigure}[t]{0.20\textwidth}
         \centering
         \includegraphics[width=\textwidth]{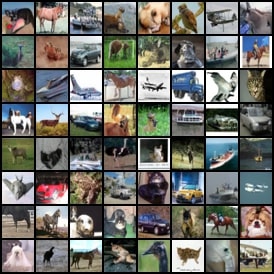}
         \caption{Progressivev2}
     \end{subfigure}
     \hspace{5mm}
     \begin{subfigure}[t]{0.20\linewidth}
         \centering
         \includegraphics[width=\linewidth]{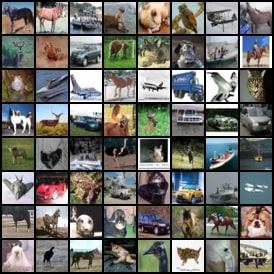}
         \caption{Progressivev3}
     \end{subfigure}  
     \\
     \caption{\textbf{Comprehensive samples visulization for unconditional diffusion model}}
     \label{appendfig:uncond_vis_all}
\end{figure}

Additionally, it's worth mentioning that for all 8x8 image grids shown in the \Cref{fig:unconditional_sample}, \ref{appendfig:uncond_vis_all}, \ref{appendfig:analysis_complete_part_1}, \ref{appendfig:analysis_complete_part_2}, \ref{appendfig:conditional_vis_1}, \ref{appendfig:conditional_vis_2}, 
\ref{appendfig:scratch_vs_partial} no matter for the unconditional diffusion model, conditional diffusion model, diffusion model for the inverse problem, or fine-tuning diffusion model, we consistently employ the same 8x8 initial noise configuration. The same setting applies to 10k initial noises for reproducibility score. This specific design is for more consistent results between different variants of diffusion models (e.g., we could clearly find the relationship between the unconditional diffusion model and conditional diffusion model by comparing \Cref{appendfig:uncond_vis_all} and \Cref{appendfig:conditional_vis_1}, \ref{appendfig:conditional_vis_2}).

\paragraph{Further discussion}

\begin{figure}[t]
     \centering
     \includegraphics[width=\linewidth]{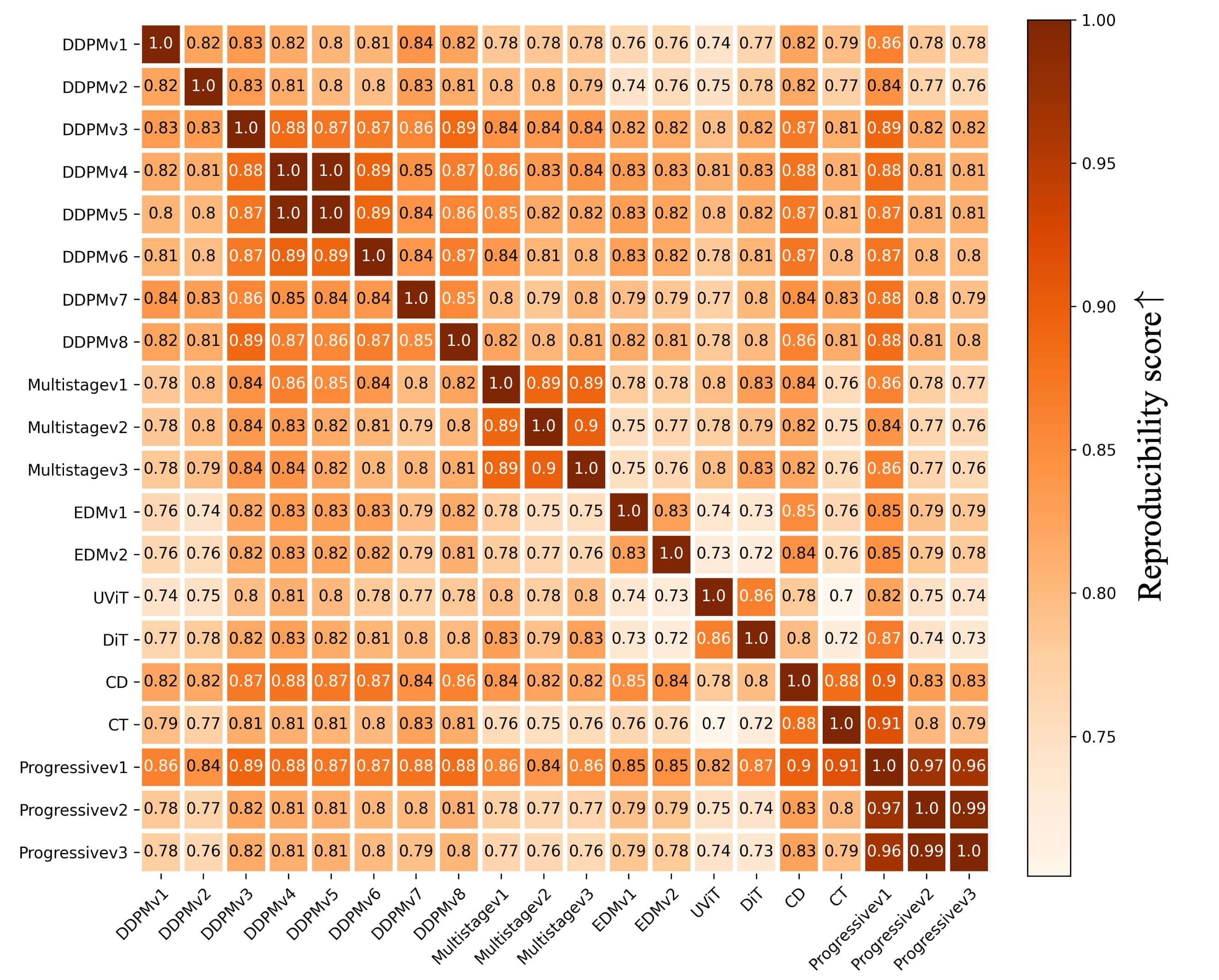}
     \caption{\textbf{Comprehensive reproducibility score among different unconditional diffusion model settings.}}
     \label{appendfig:reproducibility_sscd_all}
\end{figure}

\begin{figure}[t]
     \centering
     \includegraphics[width=\linewidth]{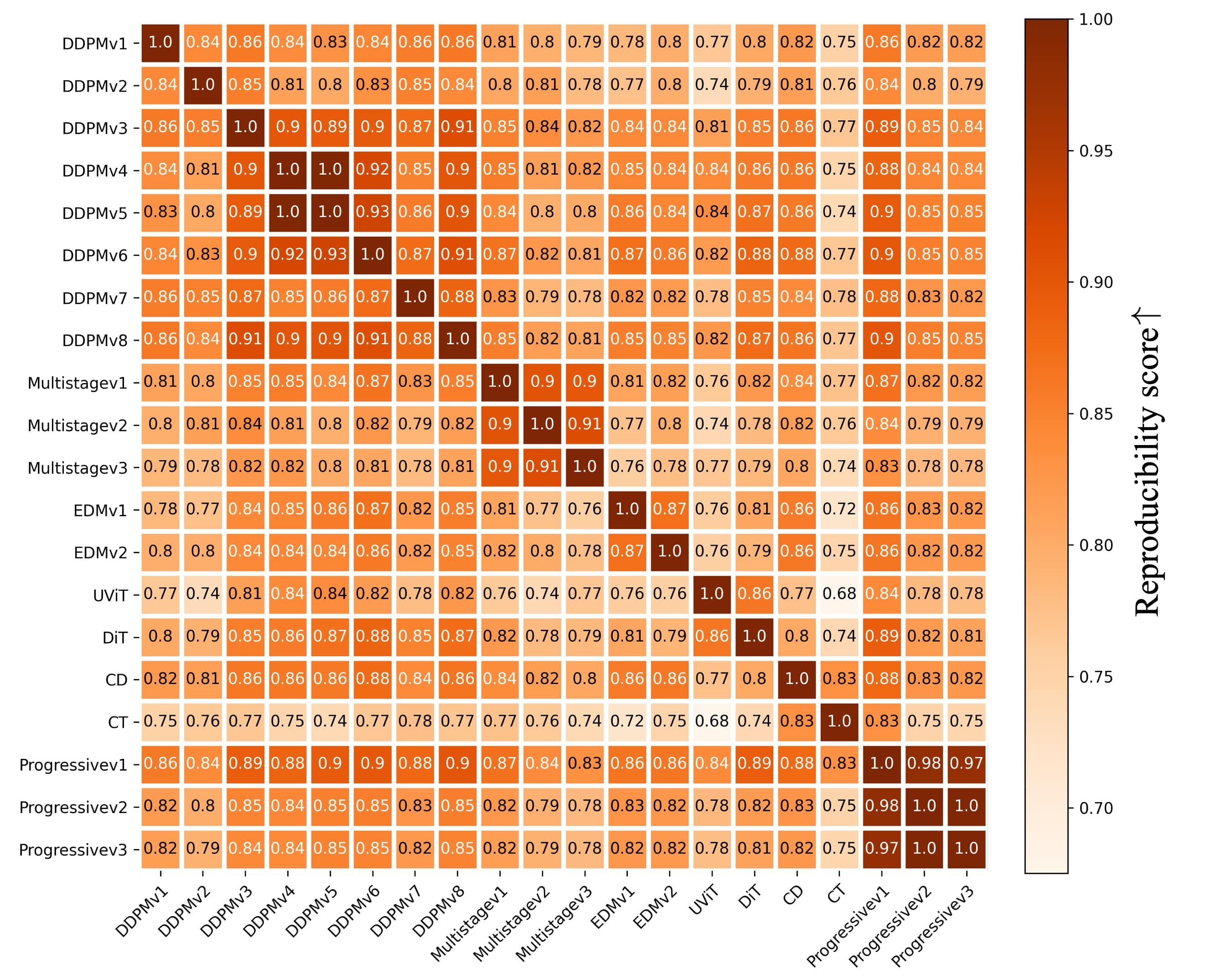}
     \caption{\textbf{Comprehensive MAE score among different unconditional diffusion model settings.}}
     \label{appendfig:reproducibility_mae_all}
\end{figure}

In \Cref{appendfig:uncond_vis_all}, we provide additional visualizations, offering a more comprehensive perspective on our findings. For a deeper understanding of our results, we present extensive quantitative data in \Cref{appendfig:reproducibility_mae_all} and \Cref{appendfig:reproducibility_sscd_all}. Building upon the conclusions drawn in \Cref{sec:analysis_generalization_unique}, we delve into the consistency of model reproducibility across discrete and continuous timestep settings. To illustrate, we compare DDPMv1 and DDPMv2, demonstrating that model reproducibility remains steadfast across these variations.Moreover, it's worth noting that while all reproducibility scores surpass a threshold of 0.6, signifying robust model reproducibility, some scores do exhibit variations. As highlighted in \Cref{appendfig:reproducibility_sscd_all}, we observe that similar architectures yield higher reproducibility scores (e.g., DDPMv1-8), models distilled from analogous teacher models exhibit enhanced reproducibility (e.g., Progressivev1-3), and models differing solely in their ODE samplers also display elevated reproducibility scores (e.g., DDPMv4, DDPMv5).We hypothesize that the disparities in reproducibility scores are primarily attributed to biases in parameter estimation. These biases may arise from factors such as differences in architecture, optimization strategies, and other variables affecting model training.
\begin{table}[t]
\begin{center}
\caption{\textbf{Comprehensive unconditional reproducibility experiment settings}}
\label{appendtab:unconditional_setting}
\begin{tabular}{llllcc}
\hline
Name & Architecture & SDE & Sampler & Continuous & Distillation \\ \hline
DDPMv1          & DDPM             & VP    & DPM-Solver        & \Checkmark &  \XSolidBrush            \\
DDPMv2          & DDPM             & VP    & DPM-Solver        & \XSolidBrush           & \XSolidBrush             \\
DDPMv3          & DDPM++             & VP    & DPM-Solver        &  \Checkmark          & \XSolidBrush             \\
DDPMv4          & DDPM++(deep)             & VP    & DPM-Solver        & \Checkmark           & \XSolidBrush             \\
DDPMv5          & DDPM++(deep)             & VP    & ODE        &  \Checkmark          &  \XSolidBrush            \\
DDPMv6          & DDPM++(deep)             & sub-VP    &  ODE       & \Checkmark           &  \XSolidBrush            \\
DDPMv7          & DDPM             & sub-VP    & ODE        &  \Checkmark          &  \XSolidBrush            \\
DDPMv8          & DDPM++             & sub-VP    & ODE        & \Checkmark           & \XSolidBrush             \\
Multistagev1    & Multistage (3 stages) & VP    & DPM-Solver        &  \Checkmark          &  \XSolidBrush            \\
Multistagev2    & Multistage (4 stages) & VP    & DPM-Solver        & \Checkmark           &  \XSolidBrush            \\
Multistagev3    & Multistage (5 stages) & VP    & DPM-Solver        & \Checkmark           &  \XSolidBrush            \\
EDMv1           & EDM             & VP    & Heun-Solver        & \Checkmark           &   \XSolidBrush           \\
EDMv2           & EDM             & VE    & Heun-Solver        & \Checkmark           &   \XSolidBrush           \\
UViT            & UViT             & VP    & DPM-Solver        & \Checkmark           &   \XSolidBrush           \\
DiT             & DiT             & VP    & DPM-Solver        &  \Checkmark          &    \XSolidBrush          \\
CD              & EDM             & VE    & 1-step         &  \Checkmark          &  \Checkmark            \\
CT              & EDM             & VE    & 1-step        &  \Checkmark          &   \XSolidBrush           \\
Progressivev1   & DDPM++          & VP    & DDIM (1-step)        &  \Checkmark          &  \Checkmark            \\
Progressivev2   & DDPM++          & VP    & DDIM (16-step)       &  \Checkmark          &  \Checkmark            \\
Progressivev3   & DDPM++          & VP    & DDIM (64-step)       &  \Checkmark          &   \Checkmark           \\ \hline
\end{tabular}
\end{center}
\end{table}

%% file: section_new/appendix/Appendix_theory.tex
\begin{figure}[t]
     \centering
     \begin{subfigure}[t]{0.6\columnwidth}
         \centering
         \includegraphics[width=\textwidth]{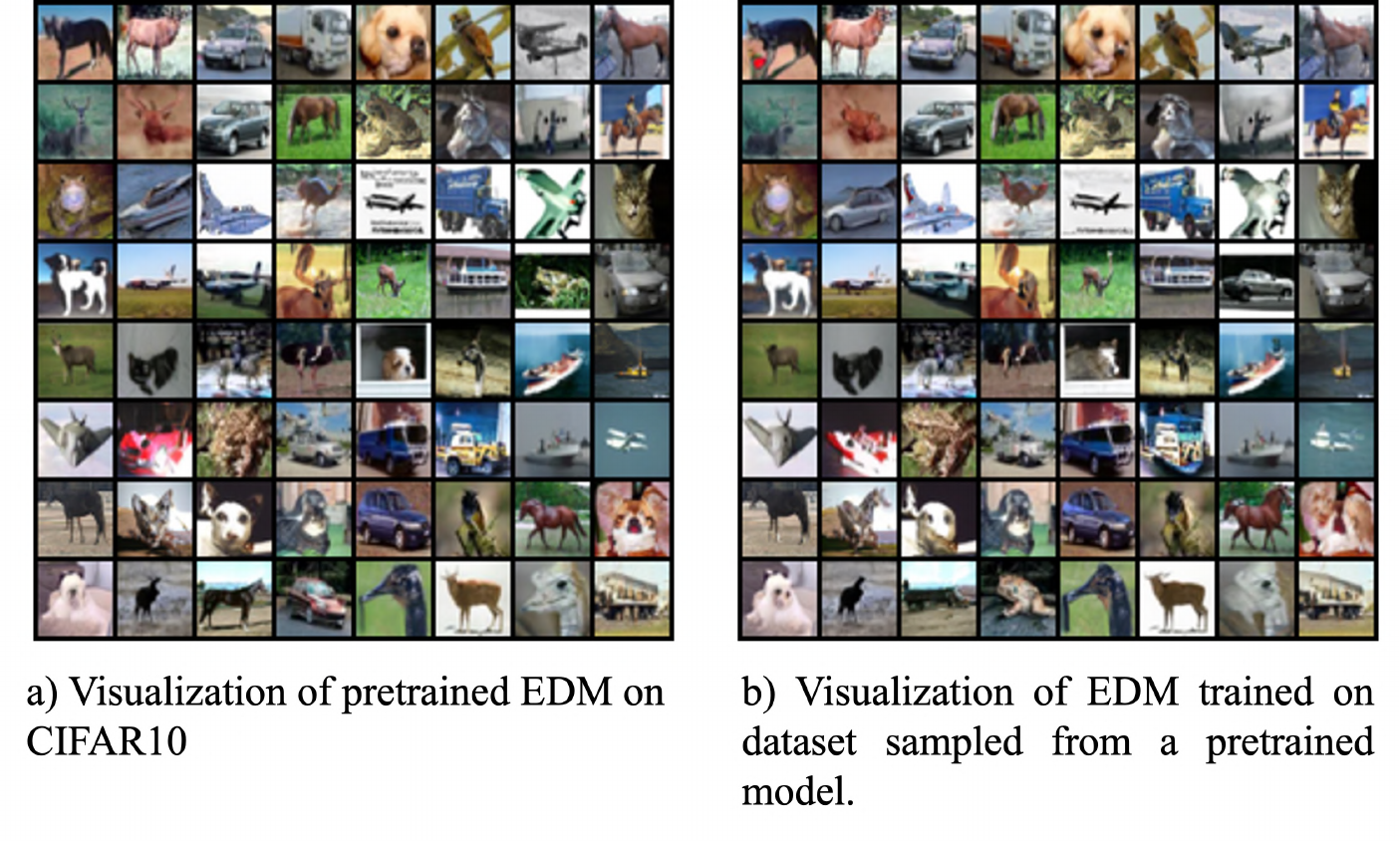}
     \end{subfigure}
     
     \caption{Pretrained model and the model trained on the sampled dataset produce almost identical results.}
     \label{fig:edm_distill_compare}
     \vspace{-0.2in}
\end{figure}

\section{Theoretical Analysis} \label{append:theory}

This section mainly focuses on the proof of \Cref{proposition:empirical distribution} in \Cref{sec:analysis_memorization}, the empirical score function would minimize the score matching loss function,  \Cref{proposition:MoG distribution} in \Cref{sec:analysis_generalization}. 


As the background, let $p_{t}(\bm x_t|\bm x_0) = \mathcal{N}(\bm x_t;s_t \bm x_0, s_t^2\sigma_t^2\textbf{I})$ be the perturbation kernel of diffusion model, which is a continuous process gradually adding noise from original image $\bm x_0$ to $\bm x_t$ along the timestep $t \in [0, 1]$. Both $s_t = s(t), \sigma_t = \sigma(t)$ here are simplified as scalar functions of $t$ to control the perturbation kernel. It has been shown that this perturbation kernel is equivalent to a stochastic differential equation $\text{d}\bm x = f(t) \bm x \text{d}t + g(t) \text{d} \bm \omega_t$, where $f(t), g(t)$ are a scalar function of $t$. The relations of $f(t), g(t)$ and $s_t, \sigma_t$ are:

\begin{equation}
    s_t = \text{exp}(\int_0^tf(\xi)\text{d}\xi), \ \ \text{and}\  \sigma_t = \sqrt{\int_{0}^{t}\frac{g^2(\xi)}{s^2(\xi)}\text{d}\xi} 
\end{equation}

\begin{mythm}{3.2}
Given a training dataset $\Brac{\bm y_i }_{i=1}^N $ of $N$-samples, consider the same setting of \Cref{lem:key} with $p(\bm x_0)$ following the empirical multi-delta distribution $p(\bm x_0) = \frac{1}{N} \sum_{i =1}^{N} \delta (\bm x_0 - \bm y_i)$. In this setting, we can show that the score function can be characterized as 
\begin{align*} 
\begin{split}
    &\bm{s}_{\text{emp}}(\bm x_t;t) = -\frac{1}{s^2_t \sigma^2_t}\brac{\bm x_t - s_t\frac{\sum_{i = 1}^{N}\mathcal{N}(\bm x_t;s_t\bm y_i, s_t^2\sigma_t^2\textbf{I})\bm y_i}{\sum_{i = 1}^{N}\mathcal{N}(\bm x_t;s_t\bm y_i, s_t^2\sigma_t^2\textbf{I})}} 
\end{split}
\end{align*}
\end{mythm}

\begin{proof}

we compute
\begin{align*}
    p_t(\bm x ) = \int p_t(\bm x | \bm x_0) p(\bm x_0) \text{d} \bm x_0 = \frac{1}{N} \sum_{i=1}^{N} \mathcal N(\bm x; s_t \bm y_i,  s_t^2 \sigma_t^2 \bm I).
\end{align*}

Therefore, the score function is:
\begin{align*}
    \bm{s}_{\text{emp}}(\bm x_t;t)
    &= \nabla_{\bm x_t} \text{log} p_t(\bm x_t ) = \frac{\nabla_{\bm x_t} p_t(\bm x_t)}{p_t(\bm x_t )} = - \frac{1}{\beta_t^2}\frac{ \sum_{i=1}^{N} \mathcal N(\bm x_t; s_t \bm y_i,  s_t^2 \sigma_t^2 \bm I) \left(\bm x_t - s_t \bm y_i\right) }{\sum_{i=1}^{N} \mathcal N(\bm x_t; s_t \bm y_i,  s_t^2 \sigma_t^2 \bm I)} \\
    &= -\frac{1}{s_t^2 \sigma_t^2} \left[\bm x_t - s_t \frac{ \sum_{i=1}^{N}  \mathcal \mathcal N(\bm x_t; s_t \bm y_i,  s_t^2 \sigma_t^2 \bm I)\bm y_i}{\sum_{i=1}^{N} \mathcal N(\bm x_t; s_t \bm y_i,  s_t^2 \sigma_t^2 \bm I)} \right] \\
\end{align*}  

From the relationship of predict $\bm \epsilon_{\text{emp}}$, predict $\bm x_{\text{emp}}$, and the score function:
\begin{align*}
    \bm \epsilon_{\text{emp}}(\bm x_t, t) 
    &= -s_t \sigma_t \bm s(\bm x_t, t) = \frac{1}{s_t \sigma_t} \left[\bm x_t - s_t\frac{ \sum_{i=1}^{N}  \mathcal N(\bm x_t; s_t \bm y_i, s_t^2 \sigma_t^2 \bm I)\bm y_i}{\sum_{i=1}^{N} \mathcal N(\bm x_t; s_t \bm y_i, s_t^2 \sigma_t^2 \bm I)} \right] \\
    \bm x_{\text{emp}}(\bm x_t, t) 
    &= \dfrac{\bm x_t - s_t \sigma_t \bm \epsilon_{\text{emp}}(\bm x_t, t)}{s_t} = \frac{ \sum_{i=1}^{N}  \mathcal N(\bm x_t; s_t \bm y_i, s_t^2 \sigma_t^2 \bm I)\bm y_i}{\sum_{i=1}^{N} \mathcal N(\bm x_t; s_t \bm y_i, s_t^2 \sigma_t^2 \bm I)} \\
\end{align*}

\end{proof}

Then given the noise prediction loss $\mathcal{L}(\bm \epsilon_{\bm\theta}; t) = \mathbb{E}_{\bm x_t \sim p_t(\bm x_t)} [|\bm \epsilon - \bm \epsilon_{\bm \theta}(\bm x_t, t)||^2]$, we will show that $\text{arg} \ \text{min}_{\bm \epsilon_{\bm\theta}(\bm x_t;t)} \mathcal{L}(\bm \epsilon_{\bm\theta};\bm x_t, t) = \bm \epsilon_{\text{emp}}(\bm x_t, t)$.

\begin{proof}



The proof is inspired from \cite{karras2022elucidating}. The loss could be calculated as:

\begin{align}
    \mathcal{L}(\bm \epsilon_{\bm\theta}; t) 
    &= \mathbb{E}_{\bm x_t \sim p_t(\bm x_t)} [|\bm \epsilon - \bm \epsilon_{\bm \theta}(\bm x_t, t)||^2] \\
    &= \int_{\mathbb{R}_d} \frac{1}{N}\sum_{i = 1}^{N} \mathcal{N}(\bm x_t;s_t\bm y_i, s_t^2\sigma_t^2\textbf{I}) ||\bm \epsilon - \bm\epsilon_{\bm\theta}(\bm x_t, t)||^2 \text{d}\bm x_t
    \label{eq:loss}
\end{align}


where $\bm \epsilon \sim \mathcal{N}(\bm 0, \textbf{I})$ is defined follow the perturbation kernel $p_{t}(\bm x_t|\bm x_0) = \mathcal{N}(\bm x_t;s_t \bm x_0, s_t^2\sigma_t^2\textbf{I})$: 

\begin{equation}
   \bm x_t = s_t\bm y_i + s_t \sigma_t \bm \epsilon \Rightarrow \bm \epsilon = \frac{\bm x_t - s_t\bm y_i}{s_t \sigma_t}
\label{eq:eps_x_relation}
\end{equation}

And $\bm \epsilon_{\bm \theta}$ is a "denoiser" network for learning the noise $\bm \epsilon$, under the assumption that the $\bm \epsilon_{\bm \theta}$ has infinite model capacity, and can approximate any continuous function to an arbitrary level of accuracy based on the Universal Approximation Theorem. So plugging Eq. \ref{eq:eps_x_relation} into \ref{eq:loss}, we could reparameterization the loss as:

\begin{equation}
    \mathcal{L}(\bm \epsilon_{\bm\theta}; t)  = \int_{\mathbb{R}_d} \underbrace{\frac{1}{N}\sum_{i = 1}^{N} \mathcal{N}(\bm x_t;s_t\bm y_i, s_t^2\sigma_t^2\textbf{I}) ||\bm \epsilon_{\bm\theta}(\bm x_t, t) - \frac{\bm x_t - s_t\bm y_i}{s_t \sigma_t}||^2}_{=:\mathcal{L}(\bm \epsilon_{\bm\theta};\bm x_t, t)} \text{d}\bm x_t 
\label{eq:loss_x}
\end{equation}

Eq. \ref{eq:loss_x} means we could minimize $\mathcal{L}(\bm \epsilon_{\bm\theta}; t)$ by minimizing $\mathcal{L}(\bm \epsilon_{\bm\theta};\bm x_t, t)$ for each $\bm x_t$. And to find the "optimal denoiser" $\bm \epsilon^*_{\bm\theta}$ that minimize the $\mathcal{L}(\bm \epsilon_{\bm\theta};\bm x_t, t)$ for every given $\bm x_t, t$:

\begin{equation}
    \bm \epsilon^*_{\bm\theta}(\bm x_t;t) = \text{arg} \ \text{min}_{\bm \epsilon_{\bm\theta}(\bm x_t;t)} \mathcal{L}(\bm \epsilon_{\bm\theta};\bm x_t, t)
\end{equation}

Since $\bm \epsilon_{\bm \theta}$ can approximate any continuous function to an arbitrary level of accuracy, this is a convex optimization problem; the solution could be solved by setting the gradient of $\mathcal{L}(\bm \epsilon_{\bm\theta};\bm x, t)$ w.r.t $\bm \epsilon_{\bm\theta}(\bm x_t;t)$ to zero:

\begin{align}
    &\nabla_{\bm \epsilon_{\bm\theta}(\bm x_t;t)} [\mathcal{L}(\bm \epsilon_{\bm\theta};\bm x_t, t)] = 0 \\
    \Rightarrow & \nabla_{\bm \epsilon_{\bm\theta}(\bm x_t;t)} [\frac{1}{N}\sum_{i = 1}^{N} \mathcal{N}(\bm x_t;s_t\bm y_i, s_t^2\sigma_t^2\textbf{I}) ||\bm \epsilon_{\bm\theta}(\bm x_t, t) - \frac{\bm x_t - s_t\bm y_i}{s_t \sigma_t}||^2] = 0 \\
    \Rightarrow & \frac{1}{N}\sum_{i = 1}^{N} \mathcal{N}(\bm x_t;s_t\bm y_i, s_t^2\sigma_t^2\textbf{I}) [\bm\epsilon^*_{\bm\theta}(\bm x;t) - \frac{\bm x_t - s_t\bm y_i}{s_t \sigma_t}] = 0 \\
    \Rightarrow & \bm\epsilon^*_{\bm\theta}(\bm x_t;t) = \frac{1}{s_t\sigma_t}[\bm x_t - s_t\frac{\sum_{i = 1}^{N}\mathcal{N}(\bm x;s_t\bm y_i, s_t^2\sigma_t^2\textbf{I})\bm y_i}{\sum_{i = 1}^{N}\mathcal{N}(\bm x_t;s_t\bm y_i, s_t^2\sigma_t^2\textbf{I})}]
    \label{eq:optim_func}
\end{align}

\end{proof}

\begin{mythm}{3.3}
Under the same setting of \Cref{lem:key} with $p(\bm x_0)$ following the MoG distribution introduced in \eqref{eqn:mlg},  we can show that the optimal score function is:
    \begin{align*}
            &\bm s_{\mathrm{MoG}}(\bm x_t, t) = \sum_{i \in [C]} \frac{\pi_i (\bm x_t, t)}{s_t^2\sigma_t^2} \left(- \bm x_t + \frac{1}{1 + \sigma_t^2} \bm U_i\bm U_i^\top \bm x_t\right),
    \end{align*}
    with  $\pi_i (\bm x_t, t) = \frac{\mathcal{N}\left(\bm x_t; \bm 0, s_t^2 \bm U_i \bm U_i^\top + s_t^2 \sigma_t^2 \bm I_d \right)}{\sum_{i \in [C]} \mathcal{N}\left(\bm x_t; \bm 0, s_t^2 \bm U_i\bm U_i^\top + s_t^2 \sigma_t^2 \bm I_d\right)}$.
\end{mythm}

\begin{proof}

First, let's consider the simplified case when $C = 1$:
    $$p(\bm x_0) = \mathcal{N}\left(\bm x_0; \bm 0, \bm U^*\bm U^{*^T}\right)$$

Which is equivalent to: 
\begin{align}
\bm x = \bm U^* \bm a,
\end{align}
where $\bm a \sim \mathcal{N}(\bm 0, \bm I_d)$.
    Then, we compute
    \begin{align*}
        p_t(\bm x_t ) 
        &= \int p_t (\bm x_t | \bm U^* \bm a) \mathcal{N}(\bm a; \bm 0, \bm I) d\bm a \\
        & = \frac{1}{(2\pi)^{n/2}s_t^n\sigma_t^n} \int \frac{1}{(2\pi)^{d/2}} \exp\left(-\frac{1}{2 s_t^2 \sigma_t^2}\|\bm x_t - s_t\bm U^*\bm a\|^2 \right)\exp\left( -\frac{\|\bm a\|^2}{2} \right)d\bm a \\
        & = \frac{1}{(2\pi)^{n/2}s_t^n\sigma_t^n} \left(\dfrac{1 + \sigma_t^2}{\sigma_t^2}\right)^{-d/2} \exp\left(-\frac{1}{2 s_t^2 \sigma_t^2 }\bm x_t^T \left( \bm I_n - \dfrac{1}{1 + \sigma_t^2 }\bm U^*\bm U^{*^T} \right) \bm x_t\right) \\
        &\quad \quad \quad \cdot \int \frac{1}{(2\pi)^{d/2}} \left(\dfrac{\sigma_t^2}{1 + \sigma_t^2}\right)^{-d/2} \exp\left(-\dfrac{1 + \sigma_t^2}{2\sigma_t^2} ||\bm a - \dfrac{1}{s_t + s_t\sigma_t^2} U^{*T} \bm x_t||_2^2\right) d\bm a \\
        & = \frac{1}{(2\pi)^{n/2}s_t^n\sigma_t^n} \left(\dfrac{1 + \sigma_t^2}{\sigma_t^2}\right)^{-d/2} \exp\left(-\frac{1}{2 s_t^2\sigma_t^2 }\bm x_t^T \left( \bm I_n - \dfrac{1}{1 + \sigma_t^2 }\bm U^*\bm U^{*^T} \right) \bm x_t\right) \\
        & = \frac{1}{(2\pi)^{n/2} \text{det} \left(s_t^2 \bm U^*\bm U^{*^T} + s_t^2 \sigma_t^2 \bm I_n \right)^{1/2}} \exp\left(-\dfrac{1}{2} \bm x_t^T \left(s_t^2\bm U^*\bm U^{*^T} + s_t^2 \sigma_t^2 \bm I_n\right)^{-1} \bm x_t\right)\\
        & = \mathcal N\left(\bm x_t; \bm 0, s_t^2 \bm U^*\bm U^{*^T} + s_t^2 \sigma_t^2 \bm I_n \right).
    \end{align*}
    Note that the fifth equality follows from 
    \begin{align*}
        &\text{det} \left(s_t^2 \bm U^*\bm U^{*^T} + s_t^2 \sigma_t^2 \bm I_n \right) = (s_t^2 + s_t^2 \sigma_t^2)^d \cdot (s_t^2 \sigma_t^2)^{n - d} \\
        &\left(s_t^2\bm U^*\bm U^{*^T} + s_t^2 \sigma_t^2 \bm I_n\right)^{-1} = \frac{1}{s_t^2 \sigma_t^2} \left( \bm I_n - \dfrac{\sigma_t^2}{1 + \sigma_t^2 }\bm U^*\bm U^{*^T} \right)
    \end{align*}
    And the score function is:
    \begin{align*}
        \bm s_{Gaussian}(\bm x_t, t) 
        &= \nabla_{\bm x_t} \text{log} p_t(\bm x_t ) = \frac{\nabla_{\bm x_t} p_t(\bm x_t)}{p_t(\bm x_t )} = - \left(s_t^2\bm U^*\bm U^{*^T} + s_t^2 \sigma_t^2 \bm I\right)^{-1} \bm x_t \\
        &= - \frac{1}{s_t^2 \sigma_t^2} \left(\bm I_d - \frac{1}{1 + \sigma_t^2} \cdot \bm U^*\bm U^{*^T}\right) \bm x_t = - \frac{1}{s_t^2 \sigma_t^2}  \bm x_t + \frac{1}{s_t^2 \sigma_t^2} \frac{1}{1 + \sigma_t^2} \bm U^*\bm U^{*^T} \bm x_t. 
    \end{align*}  

    Similarity, when the target distribution is Mixture of low rank gaussian:
    $$p(\bm x_0) = \sum_{i \in [C]}  \mathcal{N}\left(\bm x_0; \bm 0, \bm U_i^*\bm U_i^{*^T}\right)$$

    Then:

    \begin{align*}
        p_t(\bm x) 
        &= \sum_{i \in [C]} \int p_t (\bm x | \bm U_i^* \bm a) \mathcal{N}(\bm a; \bm 0, \bm I) d\bm a \\
        &= \sum_{i \in [C]} \mathcal N\left(\bm x; \bm 0, s_t^2 \bm U_i^*\bm U_i^{*^T} + s_t^2 \sigma_t^2 \bm I_n \right).
    \end{align*}

    And the score function is:
    \begin{align*}
        \bm s(\bm x, t) 
        &= \nabla_{\bm x} \text{log} p_t(\bm x ) \\
        &= \frac{\nabla_{\bm x} p_t(\bm x)}{p_t(\bm x )} \\
        &= \frac{\sum_{i}\pi_i \mathcal{N}\left(\bm x_0; \bm 0, \bm U_i^*\bm U_i^{*^T}\right) \left(- \frac{1}{s_t^2 \sigma_t^2}  \bm x + \frac{1}{s_t^2 \sigma_t^2} \frac{1}{1 + \sigma_t^2} \bm U_i^*\bm U_i^{*^T} \bm x\right)}{\sum_{i}\pi_i \mathcal{N}\left(\bm x_0; \bm 0, \bm U_i^*\bm U_i^{*^T}\right)}
    \end{align*}

\end{proof}

\begin{figure}[t]
     \centering
     \includegraphics[width=\linewidth]{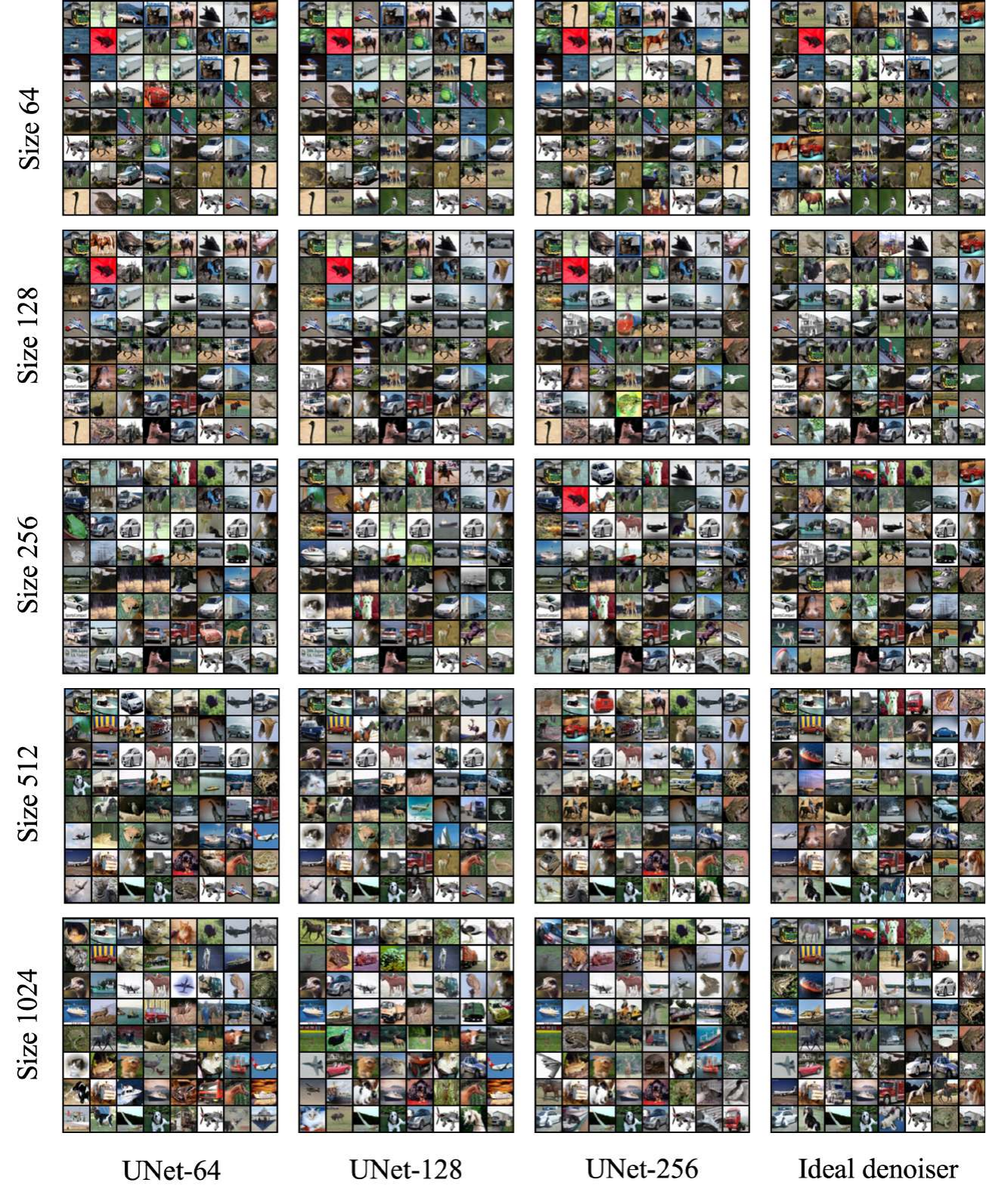}
     \caption{\textbf{Visualization between theoretical and experimental results.}}
     \label{appendfig:analysis_complete_part_1}
\end{figure}

\begin{figure}[t]
     \centering
     \includegraphics[width=\linewidth]{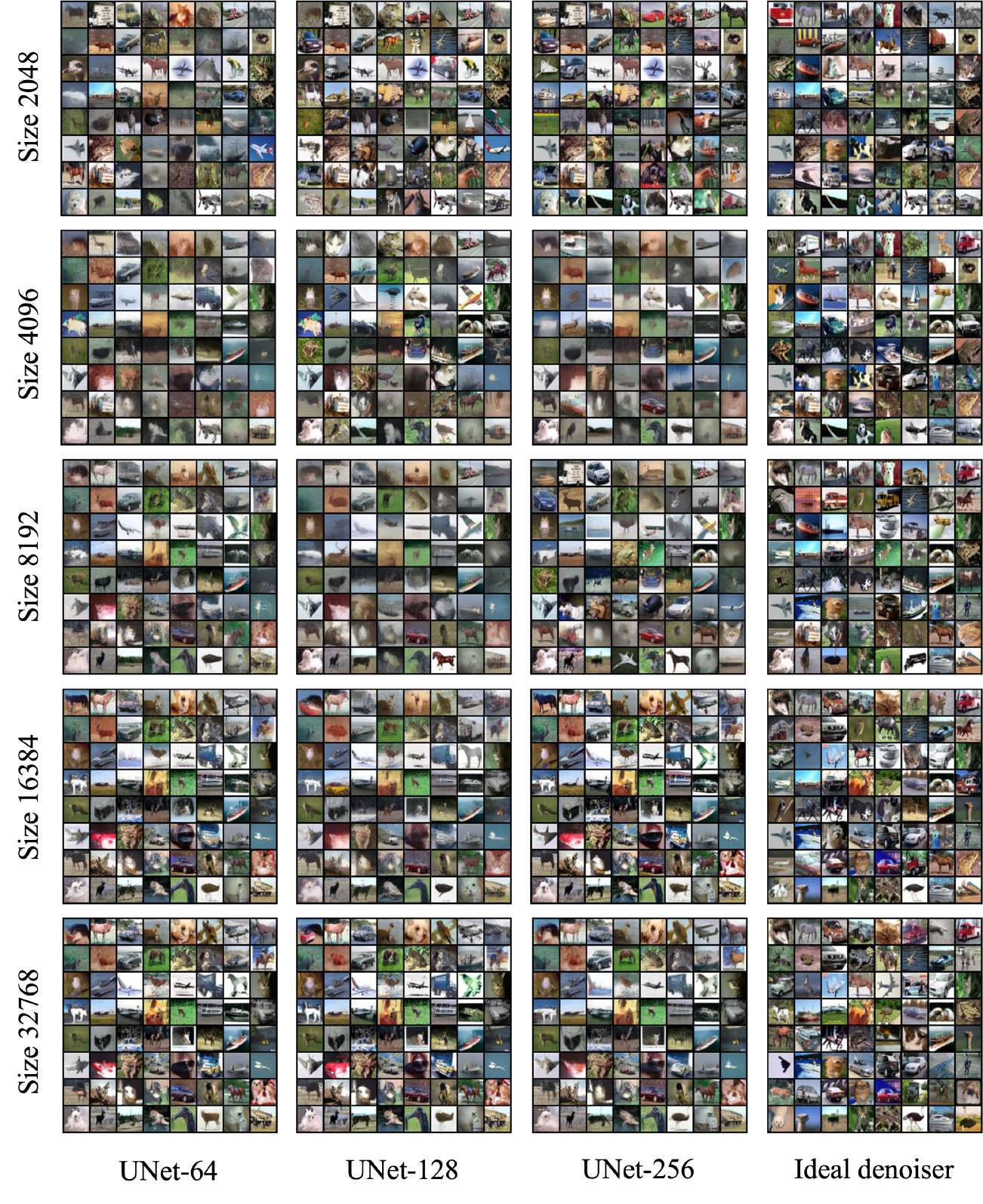}
     \caption{\textbf{Visualization between theoretical and experimental results.}}
     \label{appendfig:analysis_complete_part_2}
\end{figure}

\paragraph{Additional Experiment Setting for \Cref{fig:reproducibility_selected}}


For a more comprehensive view of our results, we present additional visualizations in  \Cref{appendfig:analysis_complete_part_1} and \Cref{appendfig:analysis_complete_part_2}. In these experiments, we train UNet models with varying numbers of channels on subsets of the CIFAR-10 dataset, each comprising different training samples. Our standard batch size for all experiments is set at 128, and we continue training until the generated samples reach visual convergence, characterized by minimal changes in both appearance and semantic information.

%% file: section_new/appendix/Appendix_distrib_learning.tex
\section{Experiment setting for Section~\ref{sec:analysis_generalization_gt}}
\label{sec:analysis_generalization_exp_setting}

\subsection{Learning score functions of a mixture of Gaussian}
\label{append:MoG}
For the mixture of Gaussian distribution, we set $C =2, d=48$. We utilize the EDM diffusion model with embed dimension 128, training with 6000 iterations for all $N$. We generate totally $100$k $(\bm x_t, t)$ pairs for estimate $\mathcal{L}_{\text{score}}$.

\subsection{Model Recovery of Diffusion Models}
\label{append:model_recovery}


In order to show how diffusion models can be recovered, we train an EDM model on the dataset sampled from a pretrained model with same architecture. We use a well-trained diffusion model in the generalization regime, the mapping of which is denoted as $f_{\bm \theta_1}$, as an implicit representation of the distribution, denoted as $p_{DM}(\bm x_0) = f_{\bm \theta_1}(\bm \epsilon), \bm \epsilon \sim \mathcal N(\bm 0, s^2_t \sigma^2_t \bm I_d)$. We sample $N$ data points $\{\bm y_i\}_{i=1}^N \subseteq \R^n$ from $p_{DM}(\bm x_0)$, following the sampling process of the diffusion model to train another diffusion model, denoted as $f_{\bm \theta_2}$. We then calculate the reproducibility of the two models $f_{\bm \theta_1}, f_{\bm \theta_2}$ following the same practice as in section \ref{sec:metric}.

In detail, $f_{\bm \theta_2}$ is pretrained on CIFAR10 and $N=50k$ which is the same as the size of CIFAR10 training set. We follow the same practive as in EDM\cite{karras2022elucidating}. We use the DDPM++ model architecure and variance preserving(VP) formulation. We train the model until convergence.

As we can see in Figure \ref{fig:edm_distill_compare}, $f_{\bm \theta_1}$ and $ f_{\bm \theta_2}$ almost generates identical results.

%% file: section_new/appendix/Appendix_conditional.tex
\section{Conditional Diffusion Models} 
\label{append:conditional}

\textbf{Extended Experiment setting} To investigate the reproducibility of the conditional diffusion model, we opted for three distinct architectures: the conditional EDM \cite{karras2022elucidating}, conditional Multistage EDM \cite{multistage}, and conditional U-ViT \cite{bao2023all}. Our training data consisted of the CIFAR-10 dataset, with the class labels serving as conditions. It's worth noting that the primary distinction between EDM and Multistage lies in the architecture of the score function. Conversely, the contrast between EDM and conditional U-ViT extends beyond architectural differences to encompass conditional embeddings. Specifically, EDM transforms class labels into one-hot vectors, subjects them to a single-layer Multilayer Perceptron (MLP), and integrates the output with timestep embeddings. In contrast, U-ViT handles class labels by embedding them through a trainable lookup table, concatenating them with other inputs, including timestep information and noisy image patches represented as tokens. For all three architectures, we pursued training until convergence was achieved, marked by the lowest FID. The DPM-Solver was employed for sampling purposes. To generate samples, we employed the same 10K initial noise distribution as utilized in the unconditional setting (refer to \Cref{sec:analysis_generalization_unique}). For each such initial noise instance, we generated 10 images, guided by 10 distinct classes, resulting in a total of 100K images.

\textbf{Discussion}  The observed reproducibility between the unconditional diffusion model and the conditional diffusion model presents an intriguing phenomenon. It appears that the conditional diffusion model learns a mapping function, denoted as $f_{c \in \mathcal C}: \mathcal E \mapsto \mathcal I_{c \in \mathcal C}$, which maps from the same noise space $\mathcal E$ to each individual image manifold $\mathcal I_{c \in \mathcal C}$ corresponding to each class $c$. In contrast, the mapping of the unconditional diffusion model, denoted as $f: \mathcal E \mapsto \mathcal I$, maps the noise space to a broader image manifold $\mathcal I \subset \bigcup_{c \in \mathcal C} \mathcal I_{c}$. A theoretical analysis of this unique reproducibility relationship holds the promise of providing valuable insights.

Currently, our research is exclusively focused on the conditional diffusion model. It raises the question of how the reproducibility phenomenon manifests in the context of the text-to-image diffusion model \cite{rombach2022high, ramesh2021zero, nichol2021glide}, where the conditioning factor is not confined to finite classes but instead involves complex text embeddings.

\begin{figure}
     \centering
     \begin{subfigure}[t]{0.20\textwidth}
         \centering
         \includegraphics[width=\textwidth]{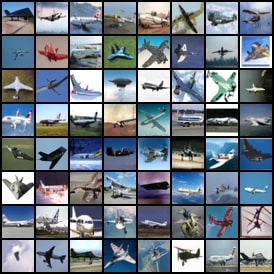}
         \caption{EDM Class0}
     \end{subfigure}
     \hspace{5mm}
     \begin{subfigure}[t]{0.20\linewidth}
         \centering
         \includegraphics[width=\linewidth]{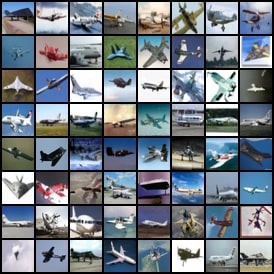}
         \caption{Multistage Class0 }
     \end{subfigure}
     \hspace{5mm}
     \begin{subfigure}[t]{0.20\linewidth}
         \centering
         \includegraphics[width=\linewidth]{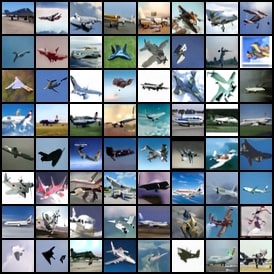}
         \caption{U-ViT Class0}
     \end{subfigure} \\
     \centering
     \begin{subfigure}[t]{0.20\textwidth}
         \centering
         \includegraphics[width=\textwidth]{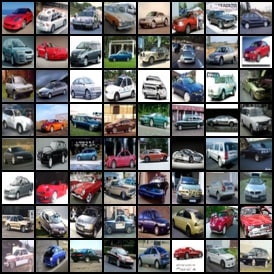}
         \caption{EDM Class1}
     \end{subfigure}
     \hspace{5mm}
     \begin{subfigure}[t]{0.20\linewidth}
         \centering
         \includegraphics[width=\linewidth]{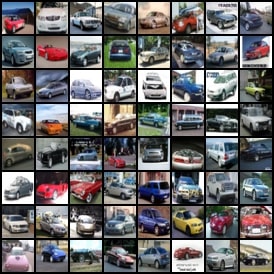}
         \caption{Multistage Class1}
     \end{subfigure}
     \hspace{5mm}
     \begin{subfigure}[t]{0.20\linewidth}
         \centering
         \includegraphics[width=\linewidth]{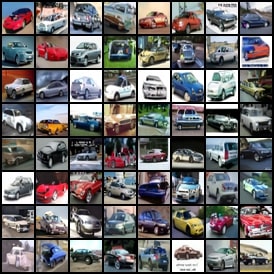}
         \caption{U-ViT Class1}
     \end{subfigure} \\
     \centering
     \begin{subfigure}[t]{0.20\textwidth}
         \centering
         \includegraphics[width=\textwidth]{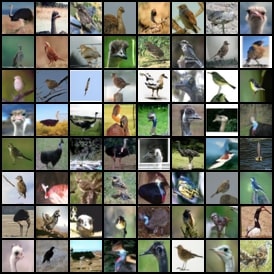}
         \caption{EDM Class2}
     \end{subfigure}
     \hspace{5mm}
     \begin{subfigure}[t]{0.20\linewidth}
         \centering
         \includegraphics[width=\linewidth]{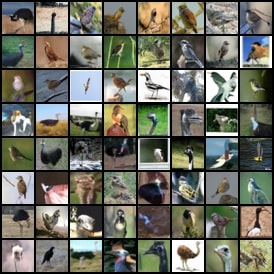}
         \caption{Multistage Class2}
     \end{subfigure}
     \hspace{5mm}
     \begin{subfigure}[t]{0.20\linewidth}
         \centering
         \includegraphics[width=\linewidth]{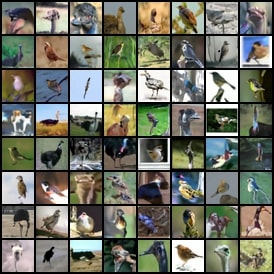}
         \caption{U-ViT Class2}
     \end{subfigure} \\
     \centering
     \begin{subfigure}[t]{0.20\textwidth}
         \centering
         \includegraphics[width=\textwidth]{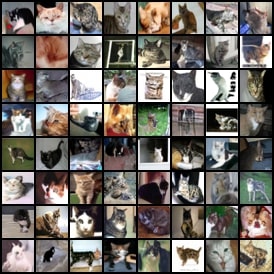}
         \caption{EDM Class3}
     \end{subfigure}
     \hspace{5mm}
     \begin{subfigure}[t]{0.20\linewidth}
         \centering
         \includegraphics[width=\linewidth]{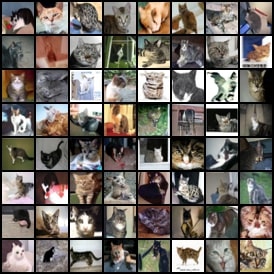}
         \caption{Multistage Class3}
     \end{subfigure}
     \hspace{5mm}
     \begin{subfigure}[t]{0.20\linewidth}
         \centering
         \includegraphics[width=\linewidth]{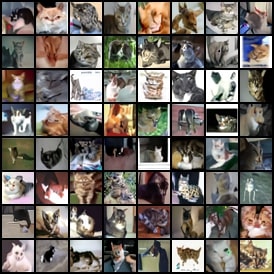}
         \caption{U-ViT Class3}
     \end{subfigure} \\
     \begin{subfigure}[t]{0.20\textwidth}
         \centering
         \includegraphics[width=\textwidth]{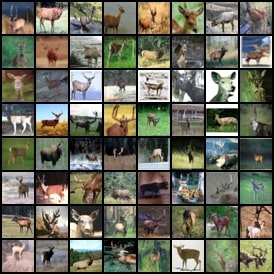}
         \caption{EDM Class4}
     \end{subfigure}
     \hspace{5mm}
     \begin{subfigure}[t]{0.20\linewidth}
         \centering
         \includegraphics[width=\linewidth]{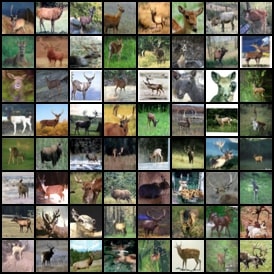}
         \caption{Multistage Class4}
     \end{subfigure}
     \hspace{5mm}
     \begin{subfigure}[t]{0.20\linewidth}
         \centering
         \includegraphics[width=\linewidth]{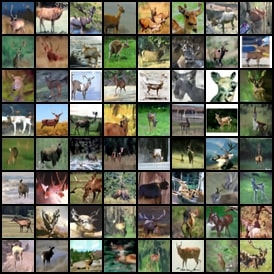}
         \caption{U-ViT Class4}
     \end{subfigure} \\ \newpage
    \vspace{0.2in}
    \caption{\textbf{Visualization of conditional diffusion model generations (class 0 - 4).}}
    \label{appendfig:conditional_vis_1}
\end{figure}

\begin{figure}[t]
     \centering
     \begin{subfigure}[t]{0.20\textwidth}
         \centering
         \includegraphics[width=\textwidth]{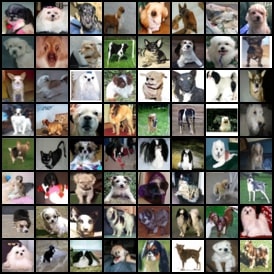}
         \caption{EDM Class5}
     \end{subfigure}
     \hspace{5mm}
     \begin{subfigure}[t]{0.20\linewidth}
         \centering
         \includegraphics[width=\linewidth]{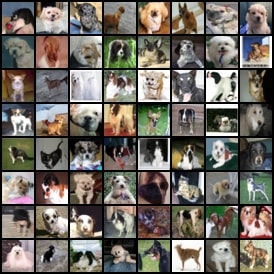}
         \caption{Multistage Class5}
     \end{subfigure}
     \hspace{5mm}
     \begin{subfigure}[t]{0.20\linewidth}
         \centering
         \includegraphics[width=\linewidth]{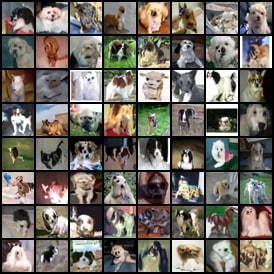}
         \caption{U-ViT Class5}
     \end{subfigure} \\
     \centering
     \begin{subfigure}[t]{0.20\textwidth}
         \centering
         \includegraphics[width=\textwidth]{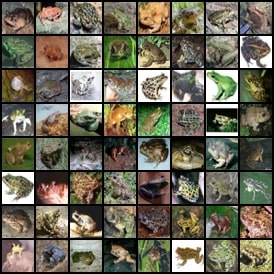}
         \caption{EDM Class6}
     \end{subfigure}
     \hspace{5mm}
     \begin{subfigure}[t]{0.20\linewidth}
         \centering
         \includegraphics[width=\linewidth]{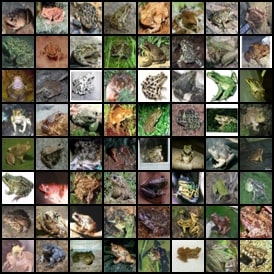}
         \caption{Multistage Class6}
     \end{subfigure}
     \hspace{5mm}
     \begin{subfigure}[t]{0.20\linewidth}
         \centering
         \includegraphics[width=\linewidth]{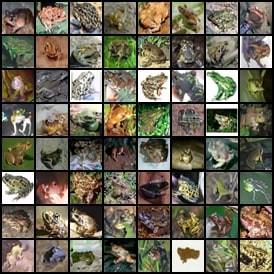}
         \caption{U-ViT Class6}
     \end{subfigure} \\
     \centering
     \begin{subfigure}[t]{0.20\textwidth}
         \centering
         \includegraphics[width=\textwidth]{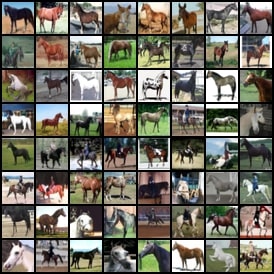}
         \caption{EDM Class7}
     \end{subfigure}
     \hspace{5mm}
     \begin{subfigure}[t]{0.20\linewidth}
         \centering
         \includegraphics[width=\linewidth]{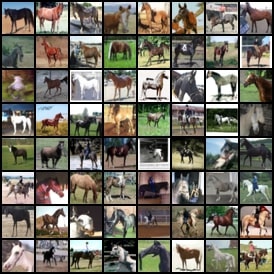}
         \caption{Multistage Class7}
     \end{subfigure}
     \hspace{5mm}
     \begin{subfigure}[t]{0.20\linewidth}
         \centering
         \includegraphics[width=\linewidth]{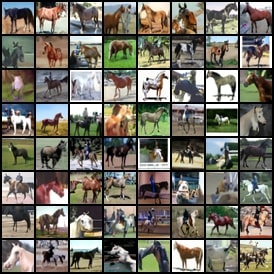}
         \caption{U-ViT Class7}
     \end{subfigure} \\
     \centering
     \begin{subfigure}[t]{0.20\textwidth}
         \centering
         \includegraphics[width=\textwidth]{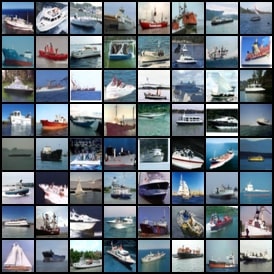}
         \caption{EDM Class8}
     \end{subfigure}
     \hspace{5mm}
     \begin{subfigure}[t]{0.20\linewidth}
         \centering
         \includegraphics[width=\linewidth]{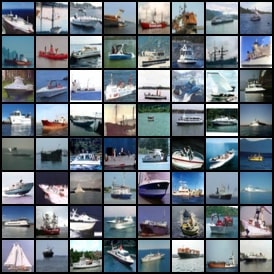}
         \caption{Multistage Class8}
     \end{subfigure}
     \hspace{5mm}
     \begin{subfigure}[t]{0.20\linewidth}
         \centering
         \includegraphics[width=\linewidth]{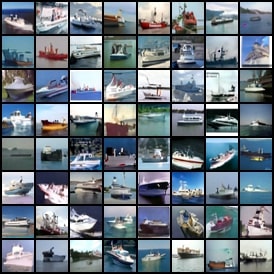}
         \caption{U-ViT Class8}
     \end{subfigure} \\
     \centering
     \begin{subfigure}[t]{0.20\textwidth}
         \centering
         \includegraphics[width=\textwidth]{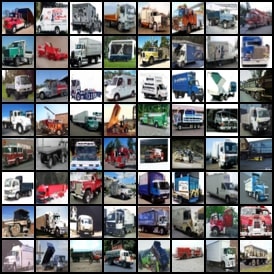}
         \caption{EDM Class9}
     \end{subfigure}
     \hspace{5mm}
     \begin{subfigure}[t]{0.20\linewidth}
         \centering
         \includegraphics[width=\linewidth]{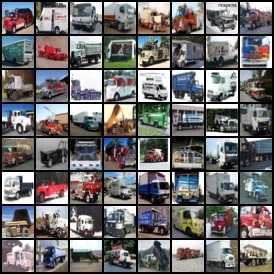}
         \caption{Multistage Class9}
     \end{subfigure}
     \hspace{5mm}
     \begin{subfigure}[t]{0.20\linewidth}
         \centering
         \includegraphics[width=\linewidth]{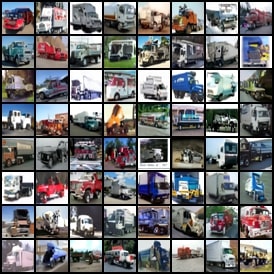}
         \caption{U-ViT Class9}
     \end{subfigure} \\     
     \caption{\textbf{Visualization of conditional diffusion model generations (class 5 - 9).}}
     \label{appendfig:conditional_vis_2}
\end{figure}

As illustrated in \Cref{appendfig:conditional_vis_1} and \Cref{appendfig:conditional_vis_2}, our previous comparisons were made with the same initial noise and class conditions. However, when comparing the same model with identical initial noise but different class conditions, we uncovered intriguing findings. For instance, the first row and column images in \Cref{appendfig:conditional_vis_1} (i) and (l) exhibited remarkable similarity in low-level structural attributes, such as color, despite differing in semantics. This observation is consistent with findings in \Cref{appendfig:scratch_vs_partial}, where we explored generation using diffusion models trained on mutually exclusive CIFAR-100 and CIFAR-10 datasets. These findings bear a striking resemblance to the conclusions drawn in \cite{khrulkov2022understanding}, which also demonstrated a similar phenomenon in a simplified scenario, where $\mathcal I$ follows a Gaussian distribution. To gain a deeper understanding of reproducibility and the phenomena mentioned in this paragraph, leveraging optimal transport methods (e.g., Schrödinger bridge \cite{shi2023diffusion, de2021diffusion, luo2023image, delbracio2023inversion, liu20232}) holds significant potential.

%% file: section_new/appendix/Appendix_stablediffusion.tex
\section{Stable Diffusion Models}
\label{append:stablediffusion}

\begin{figure}[t]
     \centering
     \begin{subfigure}[t]{0.45\textwidth}
         \centering
         \includegraphics[width=\textwidth]{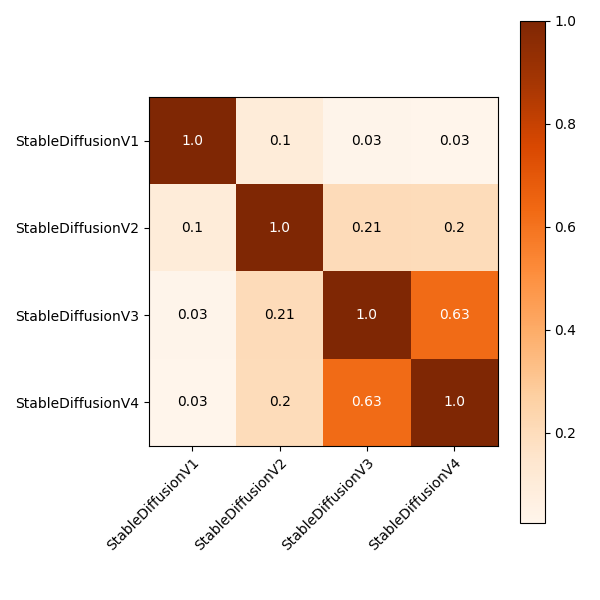}
         \caption{Reproducibility score for same initial noise}
         \label{appendfig:stablediffusion_sameinitialnoise}
     \end{subfigure}
     \hspace{5mm}
     \begin{subfigure}[t]{0.45\linewidth}
         \centering
         \includegraphics[width=\linewidth]{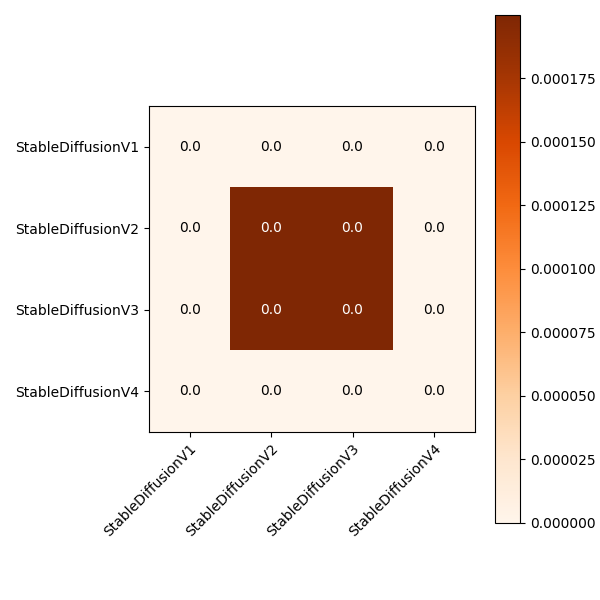}
         \caption{Reproducibility score for different initial noise}
         \label{appendfig:stablediffusion_differentinitialnoise}
     \end{subfigure}
     \begin{subfigure}[t]{1.0\linewidth}
         \centering
         \includegraphics[width=\linewidth]{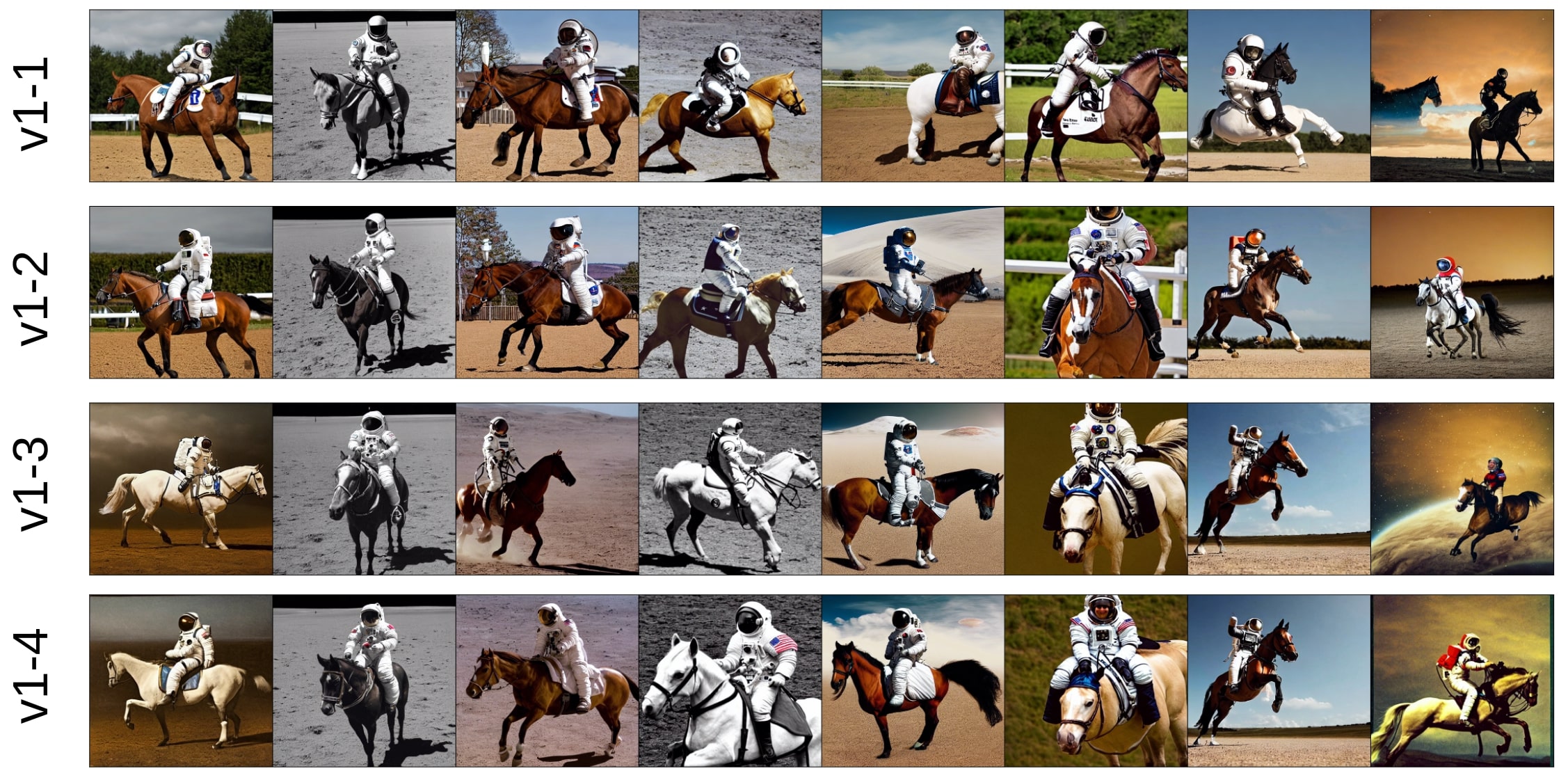}
         \caption{Visualization of stable diffusion.}
         \label{appendfig:stablediffusion_vis}
     \end{subfigure}
     \caption{\textbf{Reproducibility of Stable Diffusion.}}
     \label{appendfig:stablediffusion}
\end{figure}

Our study also explores the reproducibility of the text-to-image diffusion model, Stable Diffusion \cite{rombach2022high}, trained on the LAION-5B dataset \cite{schuhmann2022laion}. We utilize the series of pre-trained Stable Diffusion models (versions v1-1 to v1-4) released by \cite{stablediffusiongithub}. These models exhibit key differences:

\begin{itemize}
\item Versions v1-1, v1-2, and v1-3 each are trained on different subsets of the LAION-5B dataset.
\item Versions v1-3 and v1-4 share the same training subset from LAION-5B.
\item Version v1-2 is resumed from v1-1, while v1-3 and v1-4 are resumed from v1-2.
\end{itemize}

Further details on their training settings are available at \cite{stablediffusiongithub}.

For reproducibility assessment, we use the prompt "a photograph of an astronaut riding a horse" along with 1,000 randomly generated initial noises. The reproducibility score is determined with SSCD metric larger than 0.4. To isolate the impact of the guiding prompt on reproducibility, we also evaluate the reproducibility score with the same prompt but different initial noises.

The results, shown in Figure \ref{appendfig:stablediffusion_sameinitialnoise}, reveal the highest reproducibility score between v1-3 and v1-4 (0.63), likely due to their same training datasets. Lesser but noticeable reproducibility scores (below 0.21) are observed among v1-1, v1-2, and v1-3, which might be attributable to their sequential training and overlapping datasets. This finding aligns with \cite{kadkhodaie2023generalization}, suggesting that training on exclusive subsets of the same dataset can yield reproducible results in diffusion models. A notable observation in Figure \ref{appendfig:stablediffusion_vis} is the presence of flip generations between v1-3 and v1-4, potentially a result of data augmentation introducing randomness. We hypothesize that excluding data augmentation could further increase the reproducibility score between v1-3 and v1-4. Furthermore, when varying the initial noise but with the same prompt, the reproducibility scores approach zero, as evidenced in Figure \ref{appendfig:stablediffusion_differentinitialnoise}, indicating only the same prompt but different initial noise will not have reproducibility.

%% file: section_new/appendix/Appendix_inverseproblem.tex
\section{Diffusion Model for Solving Inverse Problem} \label{append:inverseproblem}

To explore the reproducibility of diffusion models in solving inverse problems, we adopted the Diffusion Posterior Sampling (DPS) strategy proposed by Chung et al. \cite{chung2022diffusion}. Our adaptation involved a slight modification of their algorithm, specifically by eliminating all sources of stochasticity within it. Additionally, we employed the DPM-Solver for Diffusion Posterior Sampling.

\textbf{Extended Experiment setting} To explore the reproducibility of diffusion models in solving inverse problems, we adopted the Diffusion Posterior Sampling (DPS) strategy proposed by Chung et al. \cite{chung2022diffusion}. Our adaptation involved a slight modification of their algorithm, specifically by eliminating all sources of stochasticity within it. Additionally, we employed the DPM-Solver for Diffusion Posterior Sampling: \Cref{alg:dps-dpm-solver}, with $N_{\text{dps}}=34$ posterior samping steps, 33 iterations for 3rd order DPM-Solver, 1 for 1st order DPM-Solver, thus 100 function evaluations. We also set all $\xi_i=1.$

\begin{figure}[t]
     \centering
     \begin{subfigure}[t]{0.9\textwidth}
         \centering
         \includegraphics[width=\textwidth]{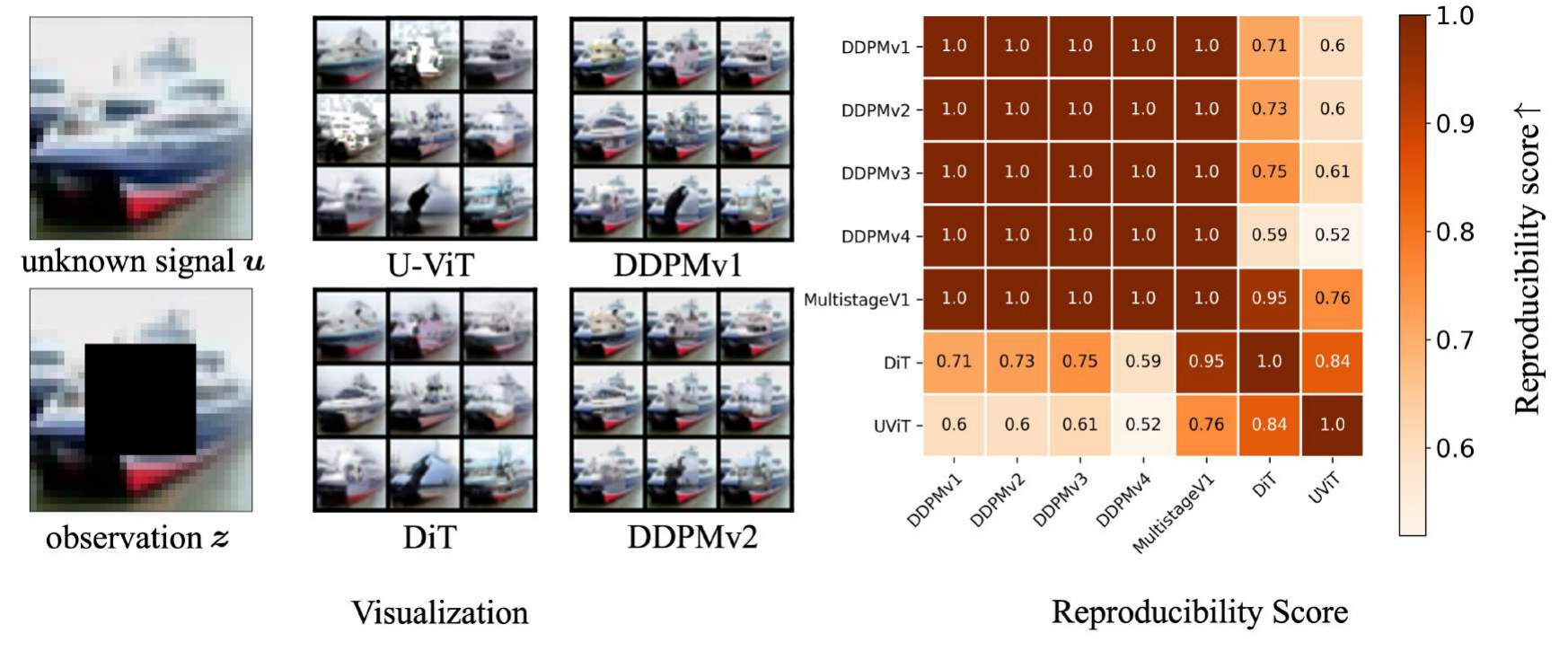}
         \caption{observation $\bm z_1$}
     \end{subfigure}\\
     \begin{subfigure}[t]{0.9\textwidth}
         \centering
         \includegraphics[width=\textwidth]{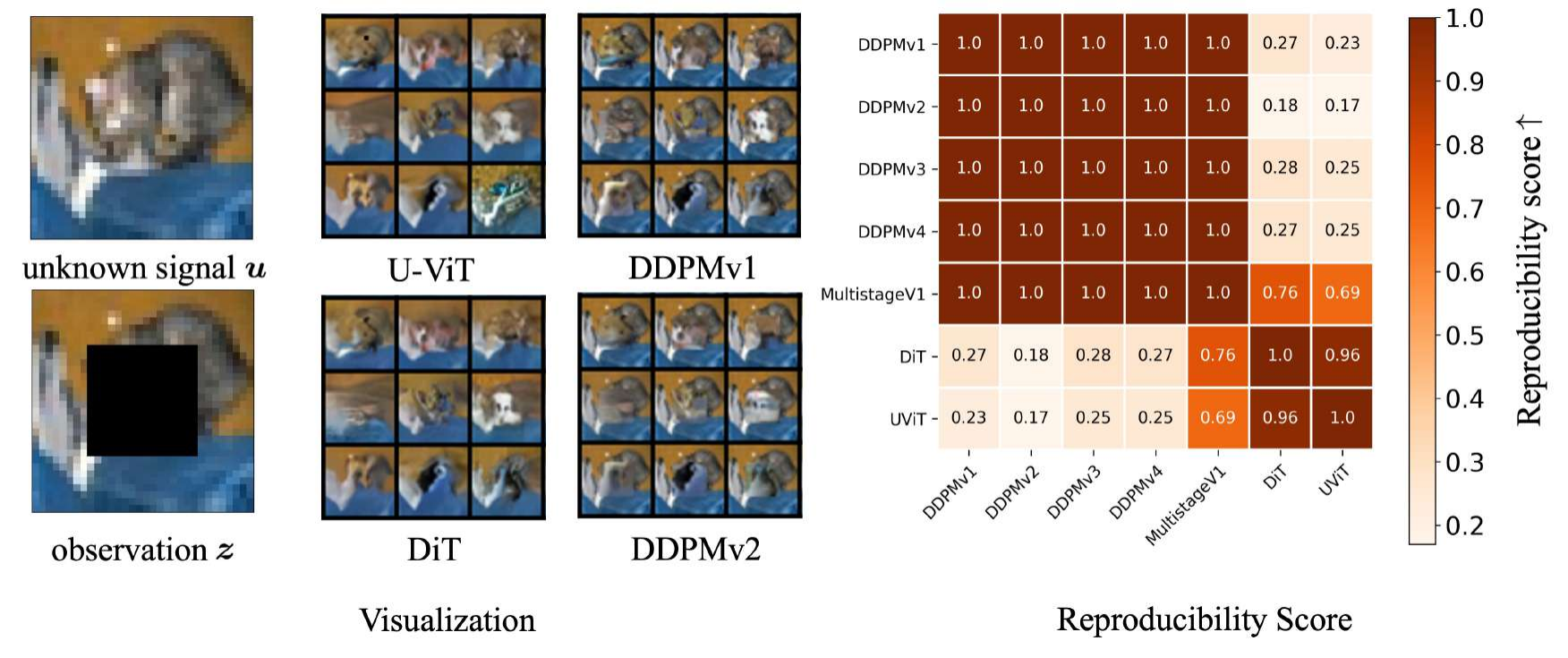}
         \caption{observation $\bm z_2$}
     \end{subfigure}\\
     \begin{subfigure}[t]{0.9\textwidth}
         \centering
         \includegraphics[width=\textwidth]{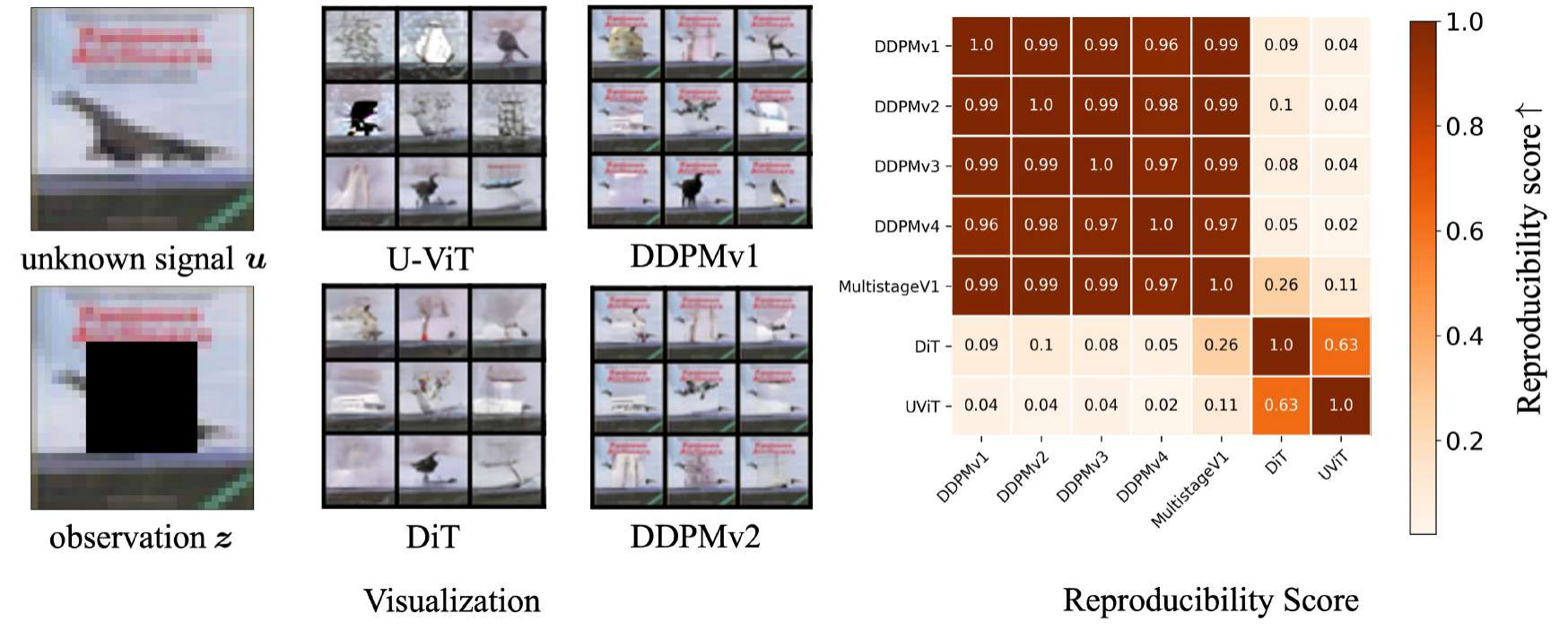}
         \caption{observation $\bm z_3$}
     \end{subfigure}\\
     \caption{\textbf{Visualization of inverse problem solving with different observations}}
     \label{appendfig:inverseproblem_analysis}
\end{figure}

\begin{figure}[htbp]
     \centering
     \includegraphics[width=1.0\linewidth]{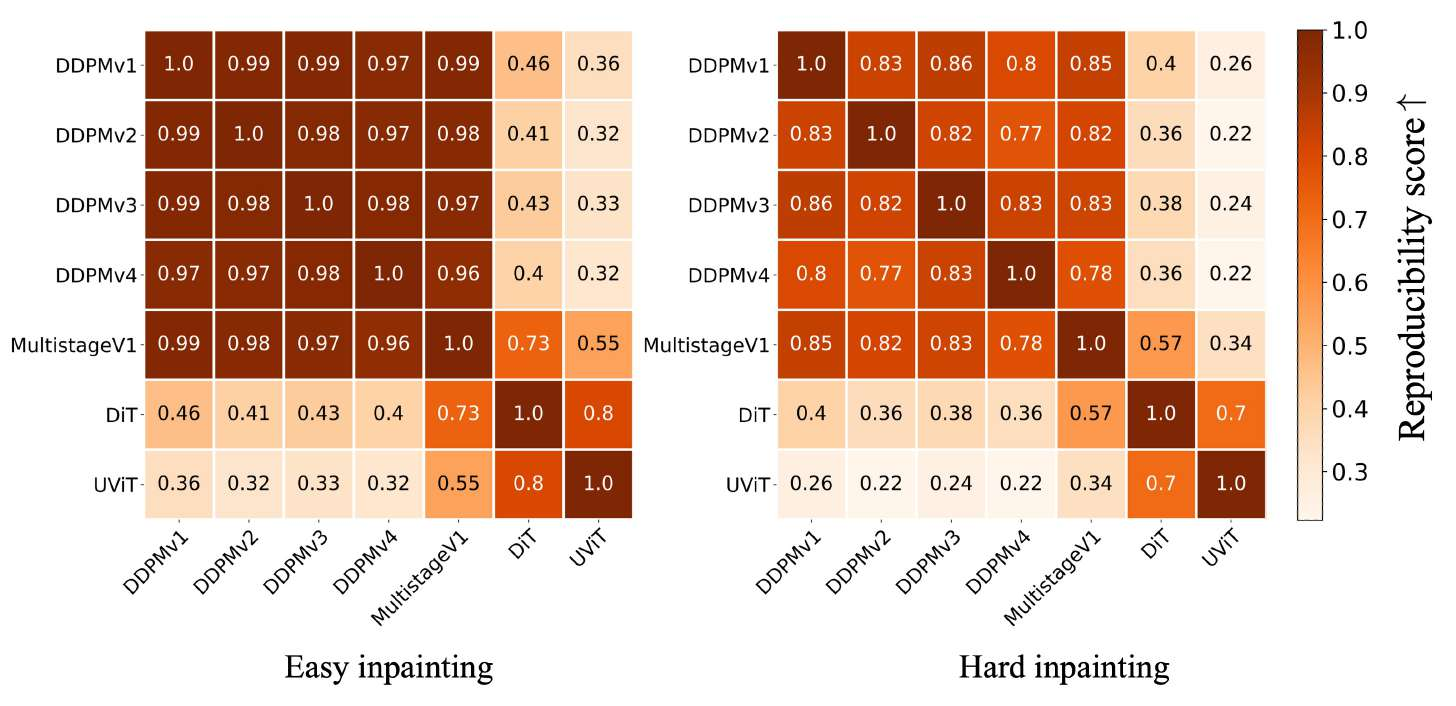}
     \caption{\textbf{Extended experiments on image impainting for reproducibility score.}}
     \label{appendfig:inverseproblem_analysis_moreobs}
\end{figure}

For the task involving image inpainting on the CIFAR-10 dataset, we applied two square masks to the center of the images. One mask measured 16 by 16 pixels, covering 25\% of the image area, and the other measured 25 by 25 pixels, covering 61\% of the image area. We denoted these as "easy inpainting" and "hard inpainting" tasks. In Figure \ref{fig:rp_inverseproblem} and Figure \ref{appendfig:inverseproblem_analysis}, we utilized the "easy inpainting" scenario with a specific observation $\bm z$ as illustrated in the figure. In Figure \ref{appendfig:inverseproblem_analysis_moreobs}, we considered both the "easy inpainting" and "hard inpainting" tasks. We also employed 10K distinct initial noise and their corresponding 10K distinct observations $\bm z$ to calculate the reproducibility score, as presented in Figure \ref{appendfig:inverseproblem_analysis_moreobs}. 
\begin{algorithm}[H]
    \centering
    \caption{Determinsitic DPS with DPM-Solver.}\label{alg:dps-dpm-solver}
    \begin{algorithmic}[1]
    \Require $N_{\text{dps}}$, $\bm u$,$f(t)$,$g(t)$, $s_t$, $\sigma_t$, \{$\xi_i\}_{i=1}^{N_{\text{dps}}} $
             \State ${\boldsymbol x}_{N_{\text{dps}}} \sim {\mathcal N}(\bm{0}, \bm{I})$
              \For{$i=N_{\text{dps}}$ {\bfseries to} $q$}
                 \State{{$\hat{\bm x}_0 = \dfrac{1}{f(i)}\paren{\bm x_i - \dfrac{g^2(i)}{s_i \sigma_i}\bm\epsilon_{\bm\theta}\paren{\bm x_i, i} }$}}
                 \State{${\boldsymbol x}'_{i-1} \gets \text{Dpm-Solver}({\boldsymbol x}_i,i)$}
                 \State{{$\bm x_{i-1} \gets \bm x'_{i-1} - \xi_i \nabla_{\bm x_i} || \bm u - \mathcal A\paren{\hat{\bm x}_0} ||^2_2$}}
              \EndFor
              \State {\bfseries return} $\hat{\mathbf{x}}_0$
    \end{algorithmic}
\end{algorithm}



\textbf{Discussion} Reproducibility is a highly desirable property when employing diffusion models to address inverse problems, particularly in contexts such as medical imaging where it ensures the reliability of generated results. As observed in \Cref{appendfig:inverseproblem_analysis}, the reproducibility scores vary for different observations $\bm z$, and the decrease in reproducibility differs across various architecture categories. For instance, when considering observation $\bm z_1$, the reproducibility scores across different architecture categories remain above 0.5, whereas for $\bm z_3$, they fall below 0.3. Since the choice of observation $\bm z$ also significantly impacts reproducibility, we conducted a complementary experiment presented in \Cref{appendfig:inverseproblem_analysis_moreobs}. In this experiment, for each initial noise instance, we employed a different observation $\bm z$. From the results, it is evident that reproducibility decreases between different categories of diffusion models. Furthermore, reproducibility diminishes as the inpainting task becomes more challenging, with "hard inpainting" being more demanding than "easy inpainting."

Here is an intuitive hypothesis of the decreasing reproducibility:

The update step of Diffusion Posterior Sampling (DPS), is constrained by the data consistency through the following equation:

\begin{equation}
    \bm x_{i-1} \leftarrow \bm x'_{i-1} - \xi_i \nabla_{\bm x_i} || \bm u - \mathcal A\paren{\hat{\bm x}_0} ||^2_2
\end{equation}

Where $\hat{\bm x}_0 = \dfrac{1}{f(i)}\paren{\bm x_i - \dfrac{g^2(i)}{s_i \sigma_i}\bm\epsilon_{\bm\theta}\paren{\bm x_i, i} }$, we could show that:

\begin{align}
    \xi_i \nabla_{\bm x_i} || \bm z - \mathcal A\paren{\hat{\bm x}_0} ||^2_2 &= \frac{\partial \mathcal A\paren{\hat{\bm x}_0}}{\partial \bm x_i}  \paren{ \mathcal A\paren{\hat{\bm x}_0} - \bm z} \\ 
    &= \frac{\partial \mathcal A\paren{\hat{\bm x}_0}}{\partial \hat{\bm x}_0} \dfrac{\partial \hat{\bm x}_0}{\partial \bm x_i} \paren{ \mathcal A\paren{\hat{\bm x}_0} - \bm z} \\
    & = \dfrac{1}{f(i)} \frac{\partial \mathcal A\paren{\hat{\bm x}_0}}{\partial \hat{\bm x}_0} \paren{1 - \dfrac{g^2(i)}{s_i \sigma_i} \dfrac{\partial \bm\epsilon_{\bm\theta} \paren{\bm x_i, i}}{ \partial x_i}} \paren{ \mathcal A\paren{\hat{\bm x}_0} - \bm z}
    \label{appendeq:inverseproblem}
\end{align}

This analysis highlights that the unconditional diffusion model is reproducible as long as the function $\bm\epsilon_{\bm\theta}$ is reproducible. However, for the diffusion model used in inverse problems to be reproducible, both the function $\bm\epsilon_{\bm\theta} \paren{\bm x_t, t}$ and its first-order derivative with respect to $\bm x_t$ must be reproducible. In other words, the denoiser should exhibit reproducibility not only in its results but also in its gradients. Combining the findings in \Cref{appendfig:inverseproblem_analysis_moreobs}, we can infer that for similar architectures, reproducibility also extends to the gradient space $\dfrac{\partial \bm\epsilon_{\bm\theta} \paren{\bm x_t, t}}{\partial \bm x_t},$ which may not hold true for dissimilar architectures. Ensuring reproducibility in the gradient space should thus be a significant focus for achieving reproducibility in diffusion models for solving inverse problems.

Additionally, it's worth noting that the data $\bm x_t$ passed into the denoiser $\bm\epsilon_{\bm\theta} \paren{\bm x_t, t}$ is always out-of-distribution (OOD) data, especially in tasks like image inpainting. Consequently, the reproducibility of OOD data $\bm x_t$ is also crucial for achieving reproducibility in diffusion models for solving inverse problems.

%% file: section_new/appendix/Appendix_finetunning.tex
\section{Fine-tuning Diffusion Model} \label{append:fintuning}

\textbf{Extended Experiment setting} In our investigation of reproducibility during fine-tuning, we first trained an unconditional diffusion model using EDM \cite{karras2022elucidating} on the CIFAR-100 dataset \cite{krizhevsky2009learning}. All the fine-tuned models discussed in this section were pre-trained on this model. Subsequently, we examined the impact of dataset size by conducting fine-tuning on the EDM using varying numbers of CIFAR-10 images: 64, 1024, 4096, 16384, and 50000, respectively. Building upon the findings in \cite{moon2022finetuning}, which indicate that fine-tuning the attention blocks is less susceptible to overfitting, we opted to target all attention layers for fine-tuning in our experiments. For comparison purposes, we also trained a diffusion model from scratch on the CIFAR-10 dataset, using the same subset of images. All models were trained for the same number of training iterations and were ensured to reach convergence, as evidenced by achieving a low Fréchet Inception Distance (FID) and maintaining consistent mappings from generated samples. The training utilized a batch size of 128 and did not involve any data augmentation.

\begin{figure}[t]
     \centering
     \includegraphics[width=1.0\linewidth]{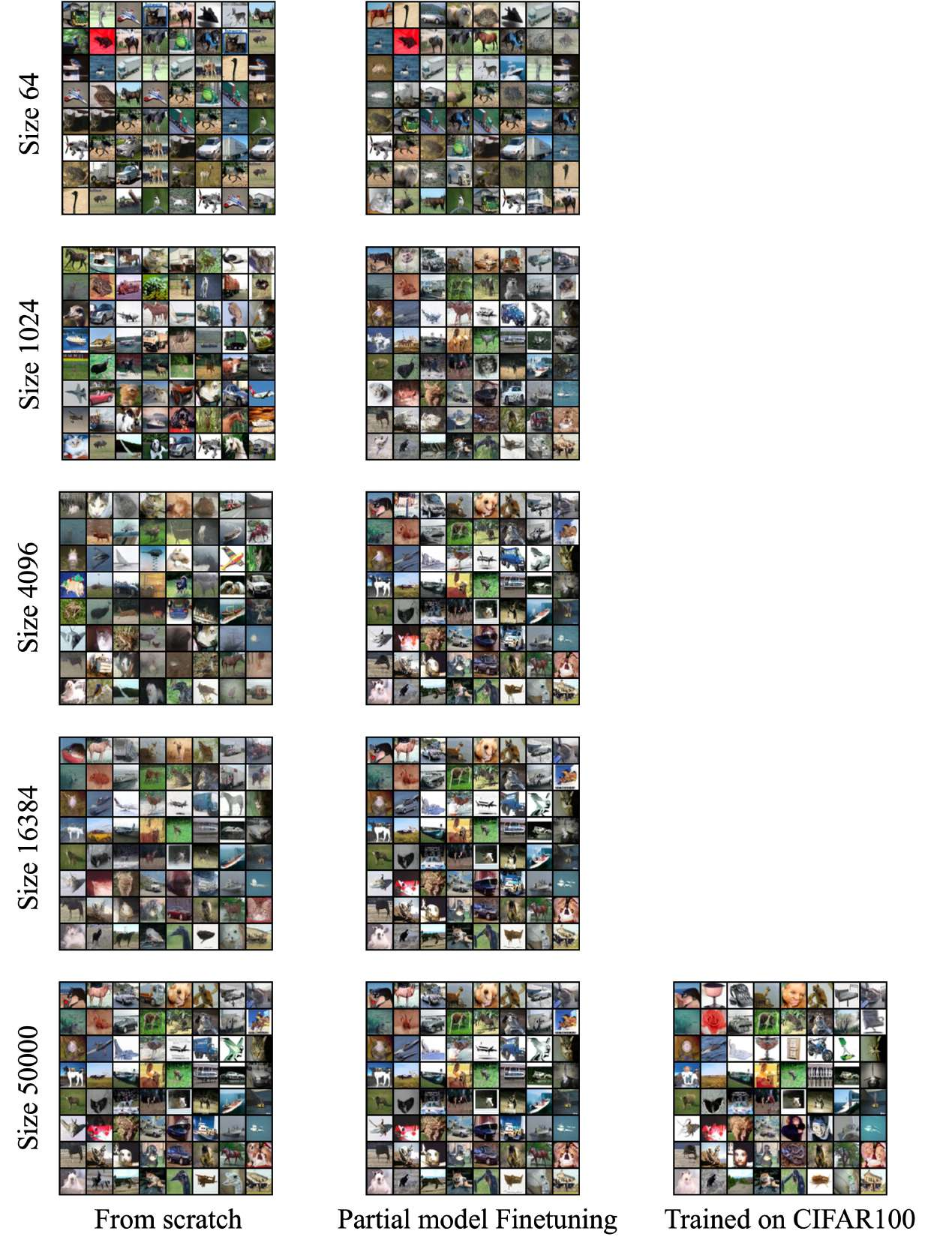}
     \caption{\textbf{More visualization of finetuning diffusion models}}
     \label{appendfig:scratch_vs_partial}
\end{figure}

\textbf{Extended Results} 
Additional generations produced by both the "from scratch" diffusion models and the fine-tuned diffusion models are presented in \Cref{appendfig:scratch_vs_partial}, encompassing various training dataset sizes. A notable observation arises when comparing the fine-tuned diffusion model's generation using 4096 and 50000 data samples. Even with this limited dataset, the fine-tuned diffusion model demonstrates a remarkable ability to approximate the target distribution. This suggests that the fixed portion of the diffusion model, containing information from the pre-trained CIFAR-100 dataset, aids the model in converging to the target distribution with less training data. In contrast, when attempting to train the diffusion model from scratch on CIFAR-10, even with 16384 data samples, it fails to converge to the target distribution. Additionally, despite the distinct nature of CIFAR-100 and CIFAR-10, their generations from the same initial noise exhibit striking similarities (\Cref{appendfig:scratch_vs_partial}). This similarity might be a contributing factor explaining how the pre-trained CIFAR-100 diffusion model assists in fine-tuning the diffusion model to converge onto the CIFAR-10 manifold with reduced training data.

